\documentclass[runningheads]{llncs}
\usepackage{graphicx}
\usepackage{amsmath,amssymb} % define this before the line numbering.
\usepackage{color}
\usepackage[width=122mm,left=12mm,paperwidth=146mm,height=193mm,top=12mm,paperheight=217mm]{geometry}
\usepackage{mathptmx}
\usepackage{graphicx}
\usepackage[hang,scriptsize]{subfigure}
\usepackage[textwidth=2cm,colorinlistoftodos]{todonotes}
\usepackage{xspace}
\usepackage{latexsym}
\usepackage{amsmath,amssymb,graphicx,xspace}
\usepackage{times}   % assumes new font selection scheme installed
\usepackage{multirow}

\usepackage{color,comment,rotating,subfigure,url,xspace} % author packages
\usepackage{tikz}
\usetikzlibrary{arrows,shadows,shapes,backgrounds,decorations,snakes,fit}
\usepackage{todonotes}
\usepackage{enumerate}
\usepackage{footnote}
\usepackage{lipsum}

\newcommand\blfootnote[1]{%
  \begingroup
  \renewcommand\thefootnote{}\footnote{#1}%
  \addtocounter{footnote}{-1}%
  \endgroup
}
\DeclareMathOperator*{\E}{\mathbb{E}}

\begin{document}
% \linenumbers
\pagestyle{headings}
\mainmatter

\title{Inter-Battery Topic Representation Learning}% Replace with your title
\titlerunning{Inter-Battery Topic Representation Learning}

\authorrunning{Cheng Zhang, Hedvig Kjellstr\"om, {Carl Henrik} Ek}

\author{Cheng Zhang$^1$, Hedvig Kjellstr\"om$^1$, Carl Henrik Ek$^2$}

\institute{{$^1$Robotics, Perception and Learning (RPL) ~~~~~~~~$^2$Department of Computer Science}\\
	KTH Royal Institute of Technology, Sweden ~~~~~~~~~~~~~~~University of Bristol,UK~~~~~~\\
	\email{~~~~ \{chengz,hedvig\}@kth.se~~~~~~carlhenrik.ek@bristol.ac.uk}
}

\maketitle

\begin{abstract}
In this paper, we present the Inter-Battery Topic Model (IBTM). Our approach extends traditional topic models by learning a factorized latent variable representation. The structured representation leads to a model that marries benefits traditionally associated with a discriminative approach, such as feature selection, with those of a generative model, such as principled regularization and ability to handle missing data. The factorization is provided by representing data in terms of aligned pairs of observations as different views. This  provides means for selecting a representation that separately models topics that exist in both views from the topics that are unique to a single view. This structured consolidation allows for efficient and robust inference and provides a compact and efficient representation. Learning is performed in a Bayesian fashion by maximizing a rigorous bound on the log-likelihood. Firstly, we illustrate the benefits of the model on a synthetic dataset,. The model is then evaluated in both uni- and multi-modality settings on two different classification tasks with off-the-shelf convolutional neural network (CNN) features which generate state-of-the-art results with extremely compact representations. 
\end{abstract}
\blfootnote{{This research has been supported by the Swedish Research Council (VR) and Stiftelsen Promobilia.}}

\section{Introduction}
\label{sec:intro}

The representation of an image has a large impact on the ease and efficiency with which prediction can be performed. This has generated a huge interest in directly learning representation from data \cite{bengio2013representation}. Generative models for representation learning treat the desired representation as an unobserved latent variable \cite{tipping1999probabilistic,blei03latent,lawrence2004gaussian}. Topic models,  which are generally a group of generative models based on Latent Dirichlet Allocation (LDA) \cite{blei03latent},  have successfully been applied for learning representations that are suitable for computer vision tasks  \cite{fei2005bayesian,hospedales11,zhang13contextual}. A topic model learns a set of topics, which are distributions over words and represents each document as a distribution over topics. In computer vision applications, a topic is a distribution over visual words, while a document is usually an image or a video. Due to its generative nature, the learned representation will provide rich information about the structure of the data with high interpretability. It offers a highly compact representation and can handle incomplete data, to a high degree, in comparison to other types of representation methodologies.  Topic models have been demonstrated with successful performance  in many applications. Similar to other latent space probabilistic models, the topic distributions can easily be adapted with different distributions with respect to the types of the input data. In this paper, we will use a LDA model as our basic framework and apply an effective factorized representation learning scheme.
\begin{figure}[t]
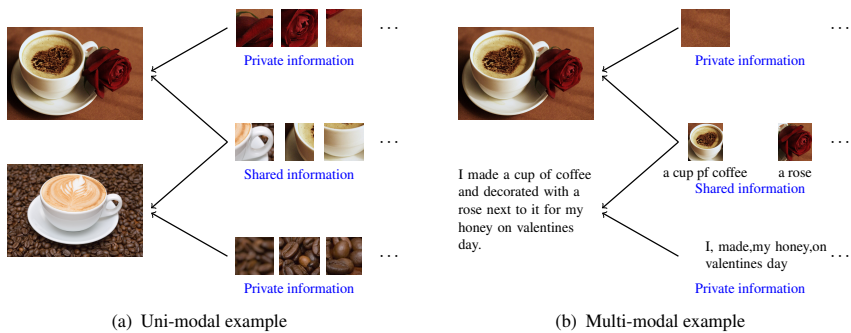

\centering
\subfigure[Uni-modal example]{
\scalebox{0.6}{\input{tikz/introPic.tex}}
}~~~
\subfigure[Multi-modal example]{
\scalebox{0.6}{\input{tikz/introPic_2.tex}}
}
\caption{ \scriptsize  Examples of using factorized representations in different scenarios. (a) gives an example of modeling "a cup of coffee" images. Different images with a cup of coffee all share certain patterns, such as cup handles, cup brims, etc. Moreover, each image also contains patterns that are not immediately related to the "cup of coffee" label, such as the rose or the coffee beans. These can be considered as private or instance-specific for each image. (b)  gives an example of modeling the image and its caption. Different modalities describe the same content as "a cup of coffee" and "a rose". However, the wooden table pattern is not described in the caption and words such  as "I made", "my honey" etc.  do not correspond to the content of the image. This information can be considered as private or modality-specific.}
\label{fig:introPic}
\end{figure}

Modeling the essence of the information among all sources of information for a particular task has been shown to offer high interpretability and better performance \cite{hospedales11,tucker58,damianou12,zhang13factorized,virtanen2012factorized,zhang14how}. For example, for object classification, separating the key features of the object from the intra-class variations and background information is key to the performance.  The idea of factorized representation can be traced back to the early work of Tucker, 'An Inter-Battery Method of Factory Analysis' \cite{tucker58}, hence, we name the model presented in this paper Inter-Battery Topic Model (IBTM).

Imagine a scenario in which we want to visually represent "a cup of coffee", illustrated in Figure \ref{fig:introPic} (a). Apart from a cup of coffee, such images commonly contain additional information that is not correlated to this labeling, e.g., the rose and the table in the upper image and the coffee beans in the lower image. One can think of the information that is common among all images of this class and thus correlated with the label, as the {\em shared} information. Images with a cup of coffee will share a set of "cup of coffee" topics between them. In addition, each image does also contain information that can be found only in a small share of the other images. This information can be thought of as {\em private}. Since the shared, but not the private, information should be employed in the estimation task (e.g., classification), it is highly beneficial to use a factorized model which represents the information needed for the tasks (shared topics) separately from the information that is not task related (private topics).

A similar idea can be applied in the case when two different modalities of the data are available. A common case is images as one modality and the captions of the images as another, as shown in Figure \ref{fig:introPic} (b). In this scenario, commonly not all of the content in the image has its corresponding caption words; and not every word in the caption has its corresponding image patches. However, the important aspects of the scene or object depicted in the image are also described in the caption, and vice versa, the central aspects of the caption are those that correlate with what is seen in the image. Based on this idea, an ideal multi-modal representation should factorize out information that is present in both modalities (words describing central concepts, and image patches from the corresponding image areas) and represent it separately from information that is only present in one of the modalities (words not correlated with the image, and image patches in the background).  Other modality examples include video and audio data captured at the same event, or optical flow and depth measurements extracted from a video stream.

To summarize, there is a strong need of modeling information in a factorized manner such that shared information and private information are represented separately. In our model, the shared part of the representation will capture the aspects of the data that are essential for the prediction (e.g., classification) task, leading to better performance. Additionally, inspecting the factorized latent representation gives a better understanding of the structure of the data, which is helpful in the design of domain-specific  modeling and data collection.

The  main contribution of this paper is
{\em a generative model, IBTM, for factorized representation learning, which efficiently factorizes essential information for an estimation task from information that is not task related} (Section \ref{sec:model}). This results in a very effective latent representation that can be used for predication tasks, such as classifications. 
IBTM is a general framework, which is applicable to both single- and multi-modal data, and can easily be  adapted to data with different noise levels. 
To infer the latent variables of the model, we derive an efficient variational inference algorithm for IBTMs.

We evaluate our model in different experimental scenarios (Section \ref{sec:exp}). Firstly, we test IBTM with a synthetic dataset to illustrate how the learning is performed. Then we apply IBTM to state-of-the-art datasets in different scenarios to illustrate how different computer vision tasks benefit from IBTM.  In a multi-modal setting, modality-specific information is factorized from cross-modality information (Section \ref{sec:ImageAndAnnot} and \ref{sec:ImageAndSegLeeds}). In a uni-modal setting,  instance-specific information is factorized from class-specific information  (Section \ref{sec:ImageAndImage} and \ref{sec:ImageAndImageLeeds}). 

\section{Related  Work}

With respect to the scope of this paper, we will summarize the related work mainly from two aspects: Topic Modeling and Factorized Models. 

\paragraph{Topic Modeling.}
Latent Dirichlet Allocation (LDA)  \cite{blei03latent} is the corner stone of topic modeling.  In computer vision tasks \cite{fei2005bayesian,hospedales11,zhang13contextual}, topic modeling assumes that each visual document is generated by selecting different themes while the themes are distributions over visual words. In correspondence with other works in representation learning, the themes can be interpreted as factors, components or dictionaries. The topic distribution for each document can be interpreted as factor weights or as a sparse and low-dimensional representation of the visual document. This has achieved  promising results in different tasks and provided an intuitive understanding of the data structure. For computer vision tasks, topic modeling has been used for classification, either with supervision in the model  \cite{blei10,zhang13c,julien08,zhu09,zhu13} or by learning  the topic representation in an unsupervised manner and applying standard classifiers such as softmax regression on the latent topic representation \cite{zhang14how}. Another interesting direction using topic modeling in computer vision is the multi-modal extension of topic models; it has been applied to tasks such as image annotation  \cite{virtanen2012factorized,wang2009simultaneous,wang2011max,blei2003modeling}, contextual action/object recognition \cite{zhang13contextual} and video tagging \cite{hospedales11}.  Being a generative model, it represents all information found in the data. However, for a specific task, only a portion of this information might be  relevant. Extracting this information is essential for a good representation of the data. Hence a model that describes key information for the current task is beneficial.  

\paragraph{Factorized Models.} The benefit of modeling  the between-view variance separately from the within-view variance was first pointed out by Tucker \cite{tucker58}. It was rediscovered in machine learning in recent years by Ek et.al.  \cite{ek2008ambiguity}. Recent research in latent structure models has also shown that modeling information in a factorized manner is advantageous for both uni-modal scenarios \cite{zhang13factorized,zhang14how,chemudugunta2006modeling}, in which only one type of data is available and multi-modal scenarios  \cite{hospedales11,damianou12,ek2008ambiguity}, in which different views correspond to different modalities. {For uni-modal scenarios, a special words topic model with a background distribution (SWB) \cite{chemudugunta2006modeling} is one of the first studies on factorized representation using topic model for information retrieval tasks. In addition to topics, SWB uses a words distribution for each document to represent document specific information and a global word distribution for background information. As shown in the experiments, this text-specific factorization model is less suitable for computer vision tasks than IBTM. } Works that apply such a factorized scheme on multi-modal topic modeling \cite{hospedales11,virtanen2012factorized} include the multi-modal factorized topic model \cite{virtanen2012factorized} and Video Tags and Topics Model (VTT) \cite{hospedales11}. The multi-modal factorized topic model which is based on correlated topic models \cite{blei2006correlated}  only provides an implicit link between different modalities with hierarchical Dirichlet priors since the factorization is enforced on the logistic normal prior,  while VTT is only designed for the specific application. 

In this paper, we present a general framework IBTM which models the topic structure in a factorized manner and can be applied to both uni- and multi-modal  scenarios. 

\section{Model}
\label{sec:model}
In this section, firstly, we will shortly review LDA \cite{blei03latent} which IBTM is based on and then present the modeling details and inference of IBTM. Finally, we will describe how the latent representation can be used for classification tasks with which we evaluate our approach. 
%%%%%
\subsection{Latent Dirichlet Allocation}
LDA is a classical generative model which is able to model the latent structure of discrete data, for example, a bag of words representation of documents.  Figure \ref{fig:FMLDA} (a) shows the graphic representation of LDA \cite{blei03latent}. In LDA, the words (visual words) $w$ are assumed to be generated by sampling from a per document topic distribution $\theta \sim Dir(\alpha)$ and a per topic words distribution $\beta \sim Dir(\sigma)$. The Dirichlet distribution is a natural choice as it is conjugate to multinomial distribution. 
%%%%%
\subsection{Inter-Battery Topic Model}
\label{sec:modelSpecification}
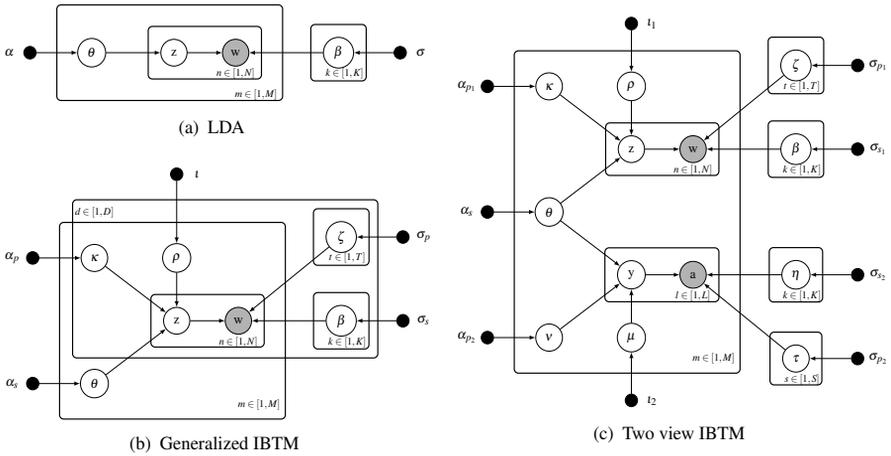
\begin{figure}[h]
\begin{minipage}{.47\textwidth}
\centering
\subfigure[LDA]{
\scalebox{0.55}{\pgfdeclarelayer{background}
\pgfdeclarelayer{foreground}
\pgfsetlayers{background,main,foreground}

\begin{tikzpicture}

\tikzstyle{surround} = [thick,draw=black,rounded corners=1mm]

% Define styles
\tikzstyle{scalarnode} = [circle, draw, fill=white!11,  
    text width=1.2em, text badly centered, inner sep=2.5pt]

\tikzstyle{scalarnodeCyan} = [circle, draw=cyan, fill=white!11,  
    text width=1.2em, text badly centered, inner sep=2.5pt]
\tikzstyle{discnode}=[rectangle,draw,fill=white!11,minimum size=0.9cm]

\tikzstyle{Vnode}=[circle, radius=1pt,draw,fill=black]
\tikzstyle{vectornode} = [circle, draw, fill=white!11,  
    text width=2.3em, text badly centered, inner sep=2pt]
\tikzstyle{state} = [rectangle, draw, text centered, fill=white, 
    text width=8em, text height=6.7em, rounded corners]

\tikzstyle{arrowline} = [draw,color=black, -latex]
\tikzstyle{carrowline} = [line width=2pt, draw,color=black, -latex]
\tikzstyle{line} = [draw]

% Nodes
\node [Vnode] at ( 0.5, 1.5) (alpha_s){};
\node [] at ( 0, 1.5) (){$\alpha$};
\node [scalarnode] at ( 2, 1.5 ) (theta) { $\theta$ };
\node [scalarnode] at ( 4, 1.5) (z) {z};
\node [scalarnode, fill=black!30] at ( 5.5, 1.5) (w) {w};
\node [scalarnode] at ( 8, 1.5) (beta) {$\beta$};
\node [Vnode] at ( 9.5, 1.5) (sigma) {};
\node [] at ( 10, 1.5) () {$\sigma$};

%Plates
\node[surround, inner sep = .3cm] (f_N) [fit = (z)(w) ] {};
\node[surround, inner sep = .5cm] (f_M) [fit = (f_N)(theta)] {};
\node[surround, inner sep = .3cm] (f_beta) [fit = (beta)] {};

% Annotate each plate
\node [] at (6, 0.5) (M) {\scriptsize $m \in [1,M]$};
\node [] at (5.5, 1) (N) {\scriptsize $n \in [1,N]$};
\node [] at (8.15, 1) () {\scriptsize $k \in [1,K]$};

% Connections 
\path [arrowline] (alpha_s) to (theta); 
\path [arrowline] (theta) to (z); 
\path [arrowline] (z) to (w); 
\path [arrowline] (beta) to (w); 
\path [arrowline] (sigma) to (beta); 

\end{tikzpicture}}
}
\subfigure[Generalized IBTM]{
\scalebox{0.55}{\pgfdeclarelayer{background}
\pgfdeclarelayer{foreground}
\pgfsetlayers{background,main,foreground}

\begin{tikzpicture}

\tikzstyle{surround} = [thick,draw=black,rounded corners=1mm]

% Define styles
\tikzstyle{scalarnode} = [circle, draw, fill=white!11,  
    text width=1.2em, text badly centered, inner sep=2.5pt]

\tikzstyle{scalarnodeCyan} = [circle, draw=cyan, fill=white!11,  
    text width=1.2em, text badly centered, inner sep=2.5pt]
\tikzstyle{discnode}=[rectangle,draw,fill=white!11,minimum size=0.9cm]

\tikzstyle{Vnode}=[circle, radius=1pt,draw,fill=black]
\tikzstyle{vectornode} = [circle, draw, fill=white!11,  
    text width=2.3em, text badly centered, inner sep=2pt]
\tikzstyle{state} = [rectangle, draw, text centered, fill=white, 
    text width=8em, text height=6.7em, rounded corners]

\tikzstyle{arrowline} = [draw,color=black, -latex]
\tikzstyle{carrowline} = [line width=2pt, draw,color=black, -latex]
\tikzstyle{line} = [draw]

% Nodes
\node [Vnode] at ( 0.5, 0) (alpha_s){};
\node [] at ( 0, 0) (){$\alpha_s$};
\node [scalarnode] at ( 2, 0 ) (theta) { $\theta$ };
\node [Vnode] at ( 0.5, 3) (alpha_p1){};
\node [] at ( 0, 3) (){$\alpha_{p}$};
\node [scalarnode] at ( 2 , 3 ) (kappa) { $\kappa$ };
\node [scalarnode] at ( 4, 1.5) (z) {z};
\node [scalarnode, fill=black!30] at ( 5.5, 1.5) (w) {w};

\node [scalarnode] at ( 4, 3) (rho) {$\rho$};
\node [Vnode] at ( 4, 5) (iota1) {};
\node [] at ( 4.5, 5) () {$\iota$};

\node [scalarnode] at ( 8, 1.5+2) (zeta) {$\zeta$};
\node [Vnode] at ( 9.5, 1.5+2) (sigma_p1) {};
\node [] at ( 10, 1.5+2) () {$\sigma_{p}$};

\node [scalarnode] at ( 8, 1.5) (beta) {$\beta$};
\node [Vnode] at ( 9.5, 1.5) (sigma_s1) {};
\node [] at ( 10, 1.5) () {$\sigma_{s}$};

%Plates
\node[surround, inner sep = .3cm] (f_N) [fit = (z)(w) ] {};
\node[surround, inner sep = .5cm] (f_M) [fit = (f_N)(theta)(rho) ] {};

\node[surround, inner sep = .3cm] (f_zeta) [fit = (zeta) ] {};
\node[surround, inner sep = .3cm] (f_beta) [fit = (beta) ] {};

\node[surround, inner sep = .2cm] (f_V) [fit = (f_N)(rho)(f_zeta)(f_beta)(kappa) ] {};
% Annotate each plate
\node [] at (6, -0.5) (M) {\scriptsize $m \in [1,M]$};
\node [] at (2, 4.1) (V) {\scriptsize $d \in [1,D]$};
\node [] at (5.5, 1) (N) {\scriptsize $n \in [1,N]$};
\node [] at (8.15, 3) () {\scriptsize $t \in [1,T]$};
\node [] at (8.15, 1) () {\scriptsize $k \in [1,K]$};

% Connections 
\path [arrowline] (alpha_s) to (theta); 
\path [arrowline] (alpha_p1) to (kappa); 
\path [arrowline] (theta) to (z); 
\path [arrowline] (kappa) to (z); 
\path [arrowline] (rho) to (z); 
\path [arrowline] (z) to (w); 
\path [arrowline] (iota1) to (rho); 

\path [arrowline] (zeta) to (w); 
\path [arrowline] (sigma_p1) to (zeta); 
\path [arrowline] (beta) to (w); 
\path [arrowline] (sigma_s1) to (beta); 
\end{tikzpicture}}
}
\end{minipage}~~~
\begin{minipage}{.47\textwidth}
\subfigure[Two view  IBTM]{
\scalebox{0.55}{\pgfdeclarelayer{background}
\pgfdeclarelayer{foreground}
\pgfsetlayers{background,main,foreground}

\begin{tikzpicture}

\tikzstyle{surround} = [thick,draw=black,rounded corners=1mm]

% Define styles
\tikzstyle{scalarnode} = [circle, draw, fill=white!11,  
    text width=1.2em, text badly centered, inner sep=2.5pt]

\tikzstyle{scalarnodeCyan} = [circle, draw=cyan, fill=white!11,  
    text width=1.2em, text badly centered, inner sep=2.5pt]
\tikzstyle{discnode}=[rectangle,draw,fill=white!11,minimum size=0.9cm]

\tikzstyle{Vnode}=[circle, radius=1pt,draw,fill=black]
\tikzstyle{vectornode} = [circle, draw, fill=white!11,  
    text width=2.3em, text badly centered, inner sep=2pt]
\tikzstyle{state} = [rectangle, draw, text centered, fill=white, 
    text width=8em, text height=6.7em, rounded corners]

\tikzstyle{arrowline} = [draw,color=black, -latex]
\tikzstyle{carrowline} = [line width=2pt, draw,color=black, -latex]
\tikzstyle{line} = [draw]

% Nodes
\node [Vnode] at ( 0.5, 0) (alpha_s){};
\node [] at ( 0, 0) (){$\alpha_s$};
\node [scalarnode] at ( 2, 0 ) (theta) { $\theta$ };
\node [Vnode] at ( 0.5, 3) (alpha_p1){};
\node [] at ( 0, 3) (){$\alpha_{p_1}$};
\node [scalarnode] at ( 2 , 3 ) (kappa) { $\kappa$ };
\node [scalarnode] at ( 4, 1.5) (z) {z};
\node [scalarnode, fill=black!30] at ( 5.5, 1.5) (w) {w};

\node [scalarnode] at ( 4, 3) (rho) {$\rho$};
\node [Vnode] at ( 4, 4.5) (iota1) {};
\node [] at ( 4.5, 4.5) () {$\iota_1$};

\node [Vnode] at ( 0.5, -3) (alpha_p2){};
\node [] at ( 0, -3) (){$\alpha_{p_2}$};
\node [scalarnode] at ( 2 , -3 ) (nu) { $\nu$ };
\node [scalarnode] at ( 4, -1.5) (y) {y};
\node [scalarnode, fill=black!30] at ( 5.5, -1.5) (a) {a};

\node [scalarnode] at ( 4, -3) (mu) {$\mu$};
\node [Vnode] at ( 4, -4.5) (iota2) {};
\node [] at ( 4.5, -4.5) () {$\iota_2$};

\node [scalarnode] at ( 8, 1.5+2) (zeta) {$\zeta$};
\node [Vnode] at ( 9.5, 1.5+2) (sigma_p1) {};
\node [] at ( 10, 1.5+2) () {$\sigma_{p_1}$};

\node [scalarnode] at ( 8, 1.5) (beta) {$\beta$};
\node [Vnode] at ( 9.5, 1.5) (sigma_s1) {};
\node [] at ( 10, 1.5) () {$\sigma_{s_1}$};

\node [scalarnode] at ( 8, -1.5) (eta) {$\eta$};
\node [Vnode] at ( 9.5, -1.5) (sigma_s2) {};
\node [] at ( 10, -1.5) () {$\sigma_{s_2}$};

\node [scalarnode] at ( 8, -1.5-2) (tau) {$\tau$};
\node [Vnode] at ( 9.5, -1.5-2) (sigma_p2) {};
\node [] at ( 10, -1.5-2) () {$\sigma_{p_2}$};
%Plates
\node[surround, inner sep = .3cm] (f_N) [fit = (z)(w) ] {};
\node[surround, inner sep = .3cm] (f_L) [fit = (y)(a) ] {};
\node[surround, inner sep = .5cm] (f_M) [fit = (f_N)(f_L)(theta)(rho)(mu) ] {};

\node[surround, inner sep = .3cm] (f_zeta) [fit = (zeta) ] {};
\node[surround, inner sep = .3cm] (f_beta) [fit = (beta) ] {};
\node[surround, inner sep = .3cm] (f_eta) [fit = (eta) ] {};
\node[surround, inner sep = .3cm] (f_tau) [fit = (tau) ] {};

% Annotate each plate
\node [] at (6, -3.5) (M) {\scriptsize $m \in [1,M]$};
\node [] at (5.5, -2) (L) {\scriptsize $l \in [1,L]$};
\node [] at (5.5, 1) (N) {\scriptsize $n \in [1,N]$};
\node [] at (8.15, 3) () {\scriptsize $t \in [1,T]$};
\node [] at (8.15, 1) () {\scriptsize $k \in [1,K]$};
\node [] at (8.15, -2) () {\scriptsize $k \in [1,K]$};
\node [] at (8.15, -4) () {\scriptsize $s \in [1,S]$};

% Connections 
\path [arrowline] (alpha_s) to (theta); 
\path [arrowline] (alpha_p1) to (kappa); 
\path [arrowline] (theta) to (z); 
\path [arrowline] (kappa) to (z); 
\path [arrowline] (rho) to (z); 
\path [arrowline] (z) to (w); 
\path [arrowline] (iota1) to (rho);

\path [arrowline] (alpha_p2) to (nu); 
\path [arrowline] (theta) to (y); 
\path [arrowline] (nu) to (y); 
\path [arrowline] (mu) to (y); 
\path [arrowline] (y) to (a); 
\path [arrowline] (iota2) to (mu); 

\path [arrowline] (zeta) to (w); 
\path [arrowline] (sigma_p1) to (zeta); 
\path [arrowline] (beta) to (w); 
\path [arrowline] (sigma_s1) to (beta); 
\path [arrowline] (eta) to (a); 
\path [arrowline] (sigma_s2) to (eta); 
\path [arrowline] (tau) to (a); 
\path [arrowline] (sigma_p2) to (tau); 
\end{tikzpicture}}
}
\end{minipage}%
\caption{Graphical representations}

\label{fig:FMLDA}
\end{figure}

We propose the IBTM which models latent variables in a factorized manner for multi-view scenarios. Firstly, we will explain how to apply IBTM to a two view scenario such that it easily can be compared to other models \cite{zhang13contextual,tucker58,wang2009simultaneous,blei2003modeling}. In the following, we present the more generalized  IBTM, which can encode any number of views.  

\paragraph{Two View IBTM.}
The two view version of IBTM, shown in Figure \ref{fig:FMLDA} (c), is an LDA-based  model, in which each document contains two views and the words $w$ and $a$ from the two views are observed respectively. The two views can represent different types of data, such as two modalities, for example, image and caption as in Figure \ref{fig:introPic} (b);  or two different descriptors for the same data, for example, SIFT and SURF features of the same image. They can also be two instances of the same class, for example, the two cups of coffee as in Figure \ref{fig:introPic} (a). 

The key of IBTM is that we assume that topics are factorized. We do not force topics from two views to be matched completely since commonly each view has its view-specific information. Hence, in our model, a shared topic distribution between two views  for each document is separated from a private topic distribution for each view. As in Figure \ref{fig:FMLDA} (c), $\theta \sim Dir(\alpha_s)$ is the {\em shared} per topic distribution for each document, and correspondingly $\beta \sim Dir(\sigma_{s1})$  and $\eta \sim Dir(\sigma_{s2})$ are the per shared topic words distributions for each view.  $\kappa \sim Dir(\alpha_{p1})$ and $\nu \sim Dir(\alpha_{p2})$ are the {\em private} per document topic distributions for each view respectively, and correspondingly $\zeta \sim Dir(\sigma_{p1})$  and $\tau \sim Dir(\sigma_{p2})$ are the private per topic word distributions for each view. To determine how much information is shared and how much information is private, partition parameters  $\rho \sim Beta(\iota_1)$ and $\mu \sim Beta(\iota_2)$ are used for each view. 
In this case, to generate topic assignments for each word in each view, $z$ and $y$ are sampled as

\begin{equation}
\scriptsize
z \sim Mult( [\rho* \theta ; (1-\rho) * \kappa]) \footnote{ \scriptsize We use $[A; B]$ to indicate matrix and vector concatenation}, ~~~~y \sim Mult([ \mu* \theta ; (1-\mu) * \nu]).
\end{equation}

In the extreme cases, if $\rho = 0$ and $\mu =0$, no information is shared between the two views and IBTM becomes two separated LDA. Otherwise,  if $\rho = 1$ and $\mu =1$, IBTM becomes a regular multi-modal topic model \cite{zhang13contextual,blei2003modeling}. 

The whole IBTM is represented as:
\begin{equation*}
\scriptsize
\begin{split}
&p(\kappa,\theta, \nu,\rho, z, w, \mu, y, a,\zeta, \beta, \eta,\tau | \Theta)\\
=& \bigg( \prod_{t=1}^{T} p(\zeta_t | \sigma_{p1} ) \bigg) \bigg( \prod_{k=1}^{K} p(\beta_k | \sigma_{s1} ) \bigg) 
 \bigg( \prod_{k=1}^{K} p(\eta_k | \sigma_{s2} ) \bigg)\bigg( \prod_{s=1}^{S} p(\tau_s | \sigma_{p2} ) \bigg)
 \prod_{m=1}^M \Bigg(p(\kappa_m | \alpha_{p_1})p(\theta_m | \alpha_{s}) \\
 &p(\nu_m | \alpha_{p_2}) p(\rho_m | \iota) p(\mu_m | \iota_2 ) 
\bigg( \prod_{n=1}^N p(z_{mn}| \kappa_m, \theta_m, \rho_{m} ) p(w_{mn} | z_{mn},\beta, \zeta )  \bigg)
\bigg( \prod_{l=1}^L p(y_{ml}| \nu_m, \theta_m, \mu_{m} )p(a_{ml} | y_{ml}, \eta, \tau)  \bigg) \Bigg)\\
\end{split}
\end{equation*}
where $\Theta = \{ \alpha_{p_1}, \alpha_{s}, \alpha_{p_2}, \sigma_{p_1}, \sigma_{p_2}, \sigma_{s_1}, \sigma_{s_2}, \iota_1, \iota_2 \} $, and as in the graphic representation of IBTM in Figure \ref{fig:FMLDA} (b), the total number of documents is $M$; the number of words for each document is $N$ and $L$ for the first view  and  the second view respectively; the number of shared topics for both views is $K$; the number of private topics  is $T$ and $S$ and the vocabulary size is $V$ and $W$ for the first view  and  the second view respectively. 
%%%%%%%%
\paragraph{Mean Field Variational Inference.}
\label{sec:VB}
Exact inference on this model is intractable due to the coupling between latent variables. Variational inference and sampling based methods are the two main groups of methods to perform approximate inference. Variational inference is known for its fast convergence and theoretical attractiveness. It can also be easily adapted to online requirements when facing big data or streaming data. Hence, in this paper, we use mean field variational inference for IBTM. The fully factorized variational distribution is assumed following the mean field manner:
\begin{equation*}
\scriptsize
\begin{split}
&q(\kappa,\theta, \nu,\rho, z, \mu, y,\zeta,  \beta, \eta,\tau)=q(\kappa) q(\theta) q(\nu) q(\rho) q(z) q(\mu) q(y) q(\zeta) q(\beta) q(\eta)q(\tau)~.
\end{split}
\end{equation*}

For each term above, the per document topic distributions are: 
$q(\kappa) = \prod_{m=1}^M q(\kappa_m | \delta_m  )$ 
where $\delta_m \in \mathbb{R}^T $; 
$
q(\theta) = \prod_{m=1}^M q(\theta_m | \gamma_m  )
$
where $\gamma_m \in \mathbb{R}^K $;
$
q(\nu) = \prod_{m=1}^M q(\nu_m | \epsilon_m  )
$
where $\epsilon_m \in \mathbb{R}^S $.
The per word topic assignments are:
$
q(z) = \prod_{m=1}^M \prod_{n=1}^N q(z_{mn} | \phi_{mn})
$
where $\phi_{mn} \in \mathbb{R}^{K+T}$  such that the first K topics correspond to the shared topics and the last T topics correspond to the private topics;
$
q(y) = \prod_{m=1}^M \prod_{l=1}^L q(y_{mn} | \chi_{mn})
$
where $\chi_{mn} \in \mathbb{R}^{K+S}$ such that the first K topics correspond to the shared topics and the last S topics correspond to the private topics. 
The per document beta parameters are:  $q(\rho) = \prod_{m=1}^M q(\rho_m | r_m)$ and $q(\mu) = \prod_{m=1}^M q(\mu_m | u_m)$. 
Finally, the per topic words distributions are: ~~$q(\zeta) = \prod_{t=1}^{T} q(\zeta_t |\xi_t  )$,  ~~$q(\beta) = \prod_{k=1}^{K} q(\beta_k | \lambda_k  )$, ~~
$q(\eta) = \prod_{k=1}^{K} q(\eta_k | \upsilon_k  )$,  ~~
$q(\tau) = \\\prod_{s=1}^{S}  q(\tau_s | \omicron_s   )$. All the variational distributions follow the same family of distributions under the model assumption. 

Applying Jensen's inequality on the log likelihood of the model, we get the evidence lower bound (ELBO) $\mathcal{L}$:
\begin{equation*}
\scriptsize
\begin{split}
&\log p( w, a, \mathbb{Z} | \Theta)= \log \int  \frac{ p(w, a, \mathbb{Z}| \Theta ) q(\mathbb{Z})  }{q(\mathbb{Z}) } d \mathbb{Z} 
 \ge \E _q[\log p(w,a,\mathbb{Z}| \Theta )] - \E_{q}[\log q(\mathbb{Z} ) ] = \mathcal{L}
\end{split}
\end{equation*}
where $\mathbb{Z}=\{ \kappa,\theta, \nu,\rho, z,  \mu, y,  \zeta,\beta, \eta,\tau\} $.

By maximizing the ELBO, we get the update equations for the variational parameters. Only the ones that differ from LDA are presented here and derivation details are presented in the supplementary material.  
The update equations for the per document topic variational distribution are:

{\scriptsize
\begin{equation*}
\delta_{mt} = \alpha_{p1} + \sum_{n=1}^N  \phi_{mn(K+t) },~~~
\gamma_{mk}=\alpha_{s}+ \sum_{n=1}^N \phi_{mnk} +\sum_{l=1}^L \chi_{mlk},~~~
\epsilon_{ms} = \alpha_{p2} +\sum_{l=1}^L \chi_{ml(K+s)}.
\end{equation*}}

The update equation for the topic assignment in the first view is, when $i\le K$:
\begin{savenotes}
\begin{equation*}
\scriptsize
\begin{split}
\phi_{mni}&=\exp\Bigg( \Big( \Psi(\gamma_{mk} ) - \Psi(\sum_{i=1}^K \gamma_{mi} )  \Big) 
+ \Big(\Psi(r_{m1} ) - \Psi(r_{m1} + r_{m2} )  \Big)
 + \sum_{v=1}^V {[w_{mn}=v]} \left( \Psi(\lambda_{iv})- \Psi(\sum_{p=1}^V \lambda_{ip} ) )  \right) -1 \Bigg);\footnote{\scriptsize $\Psi (x) $ is the digamma function.}
\end{split}
\end{equation*}
\end{savenotes}

and when $i > K$  (as $i=K+t$):
\begin{equation*}
\scriptsize
\begin{split}
\phi_{mni}&=\exp\Bigg(  \Bigg( \Big( \Psi(\delta_{m(i-K)} ) - \Psi(\sum_{p=1}^T \delta_{mp} )  \Big) 
+ \Big( \Psi(r_{m2} ) - \Psi(r_{m1} + r_{m2} )\Big) \Bigg)
+  \sum_{v=1}^V{[w_{mn}=v]}\left( \Psi(\xi_{iv})- \Psi(\sum_{p=1}^V \xi_{ip} ) )  \right)-1
\Bigg)~.
\end{split}
\end{equation*}

The update equations for the partition parameters are:
{\scriptsize
\begin{equation*}
r_{m1}=  \iota_{11} +\sum_{n=1}^N \sum_{i=1}^K \phi_{mni},~~~~
r_{m2}=  \iota_{12} +\sum_{n=1}^N \sum_{i=K}^{K+T} \phi_{mni}
\end{equation*}}
The update for the second view follows equivalently. 

In the implementation, all global latent variables are initialized randomly except for the shared per topic word distribution for the second modality, which is initialized uniformly. Due to the exchangeability of Dirichlet distribution which leads to rotational symmetry in the inference, initializing only one of the shared per topic word distribution randomly will increase the robustness of the model performance. 
%\footnote{The code of the model will be released after the paper being published. }
%%%%%%%%%

\paragraph{Generalized  IBTM.}
It is straight-forward to generalize the two view IBTM to more views. The graphical representation of the generalized IBTM is shown in Figure \ref{fig:FMLDA} (b), where $D$ is the total number of views. When $D=2$, the models in Figure \ref{fig:FMLDA} (b) and \ref{fig:FMLDA} (c) are identical. The inference procedure can be adapted easily,  since the updates of both topic assignments and partition parameters for each view follow the same form. The only difference is the per document shared topic variational distribution $\gamma_{mk}=\alpha_{s}+ \sum_{d=1}^D  \sum_{n=1}^{N^{(d)}} \phi_{mnk}^{(d)} $, where  $\phi_{mnk}^{(d)}$ is the variational distribution of the topic assignment for the $d$-th view. 
%%%%%%%%%%%%%%%

\subsection{Classification}
Topic models provide a compact representation of the data. Both LDA and IBTM are unsupervised models and can be used for representation learning.  The topic representation can be applied to different tasks, for example, image classification and image retrieval. Commonly, the whole topic representation will be employed for these tasks using LDA. Using IBTM, we will only rely on the shared topic space which represents the information essence.  
For image classification, we can simply apply a Support Vector Machine or softmax regression, taking the shared topic representation as the input. In our experimental evaluation, softmax regression is used. 
Although there are different types of supervised  topic models \cite{blei10,zhu09} where class label is encoded as part of the model, the work in \cite{zhang14how} shows that the performance  on computer vision classification tasks using supervised model and unsupervised model with an additional classifier is similar. The minor improvement on the performance commonly comes with significant improvement of computation cost. Hence, we keep IBTM as a general framework for representation learning in an unsupervised manner.

\setcounter{secnumdepth}{6}
\section{Experiments}
\label{sec:exp}
In the experiments, firstly, we will evaluate the inference  scheme and demonstrate the model behavior in a controlled manner in Section \ref{sec:exp_syn}. Then we will use two benchmark datasets to evaluate the model behavior in real world scenarios in Section \ref{sec:exp_bm}. For this purpose, we use the LabelMe natural scene data for natural scene classification \cite{wang2009simultaneous,li2012objects,zheng14topic}  and the Leeds butterfly dataset \cite{wang2009Learning} for fine-grained categorization. 

%%%%%%%%
\begin{figure}[t]
\centering
\begin{minipage}{.45\textwidth}
\subfigure[Ground Truth  $\zeta,\beta,\eta,\tau$]{
\includegraphics[width=1.3cm]{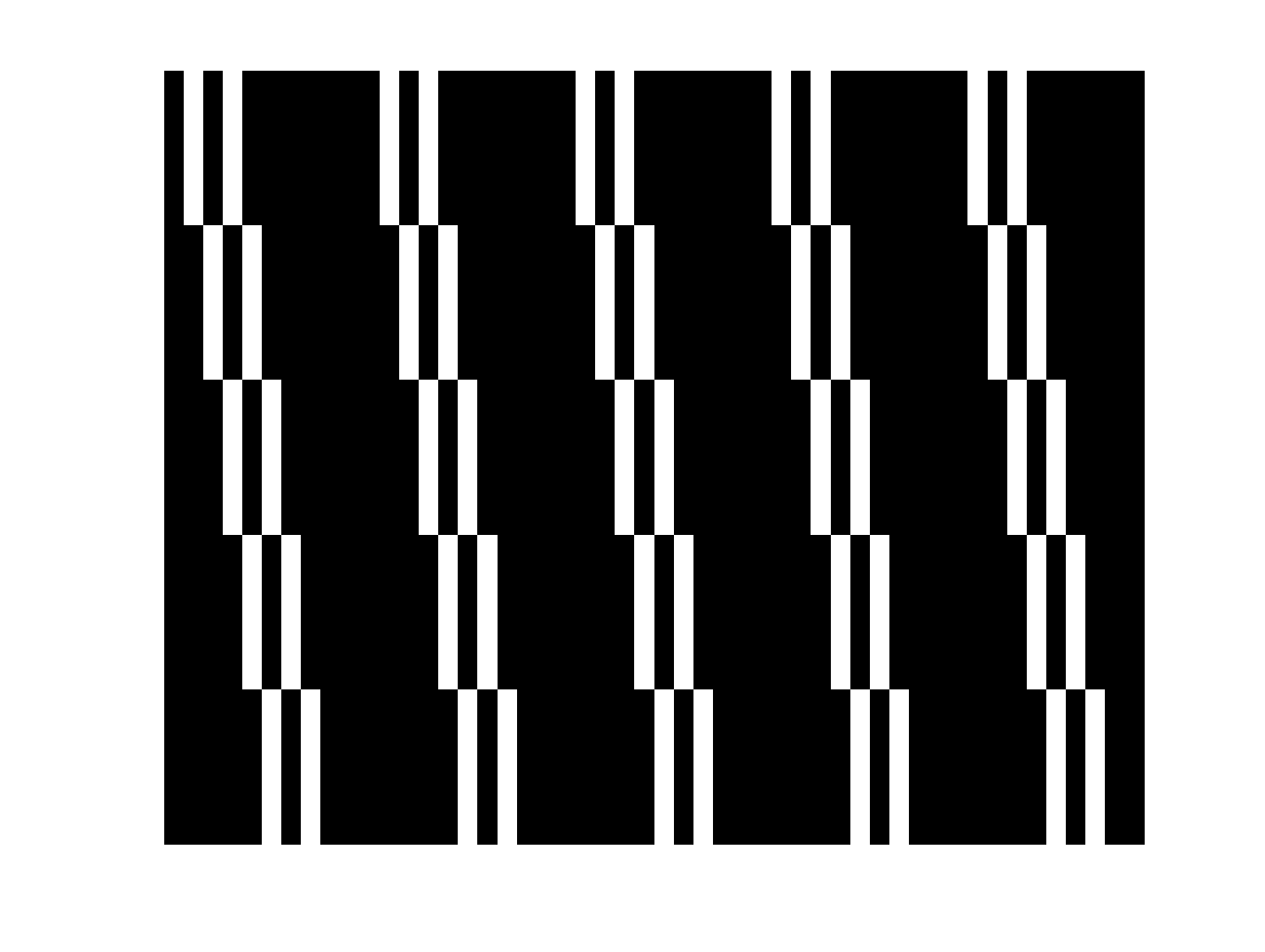}
\includegraphics[width=1.3cm]{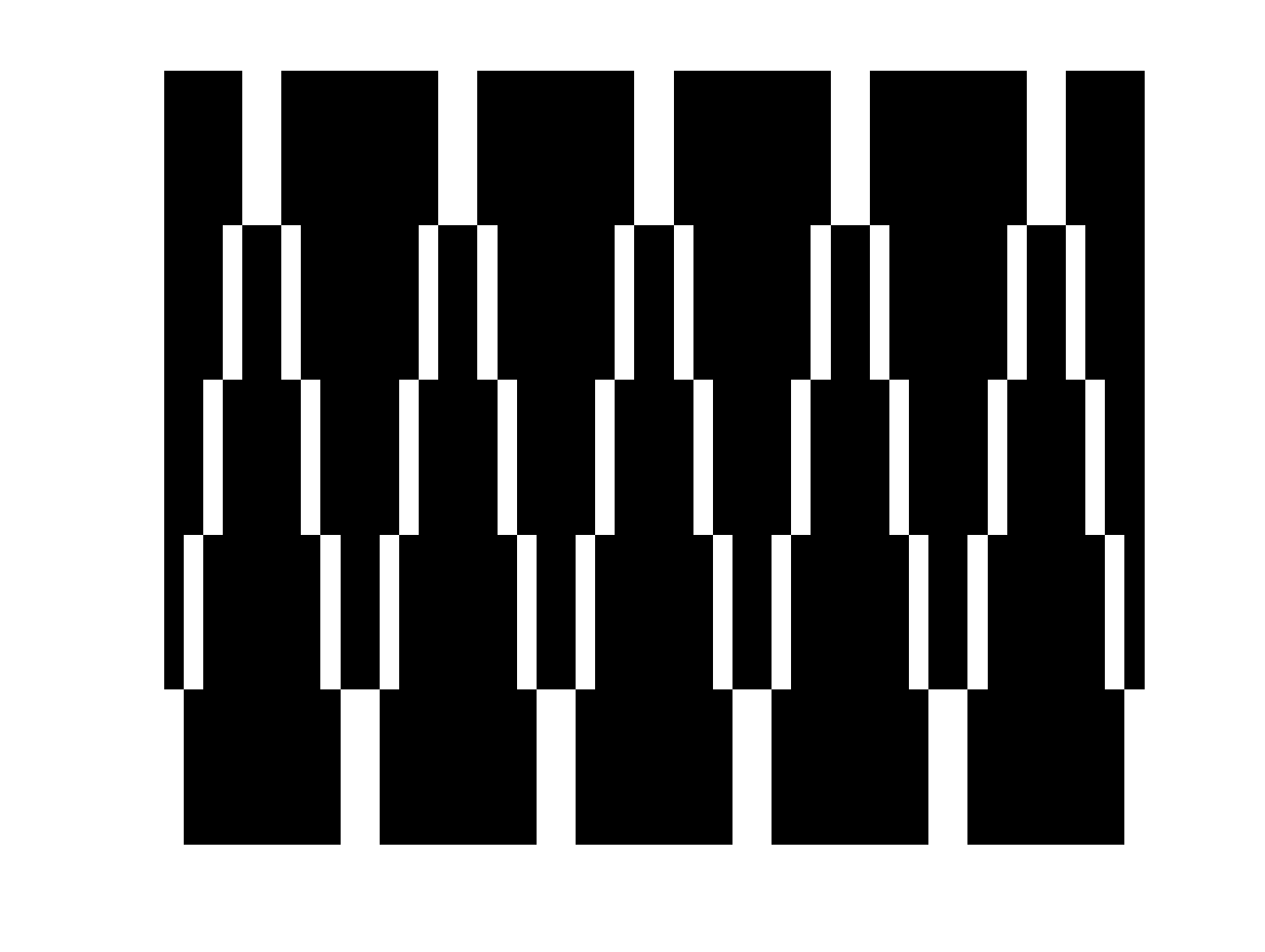}
\includegraphics[width=1.3cm]{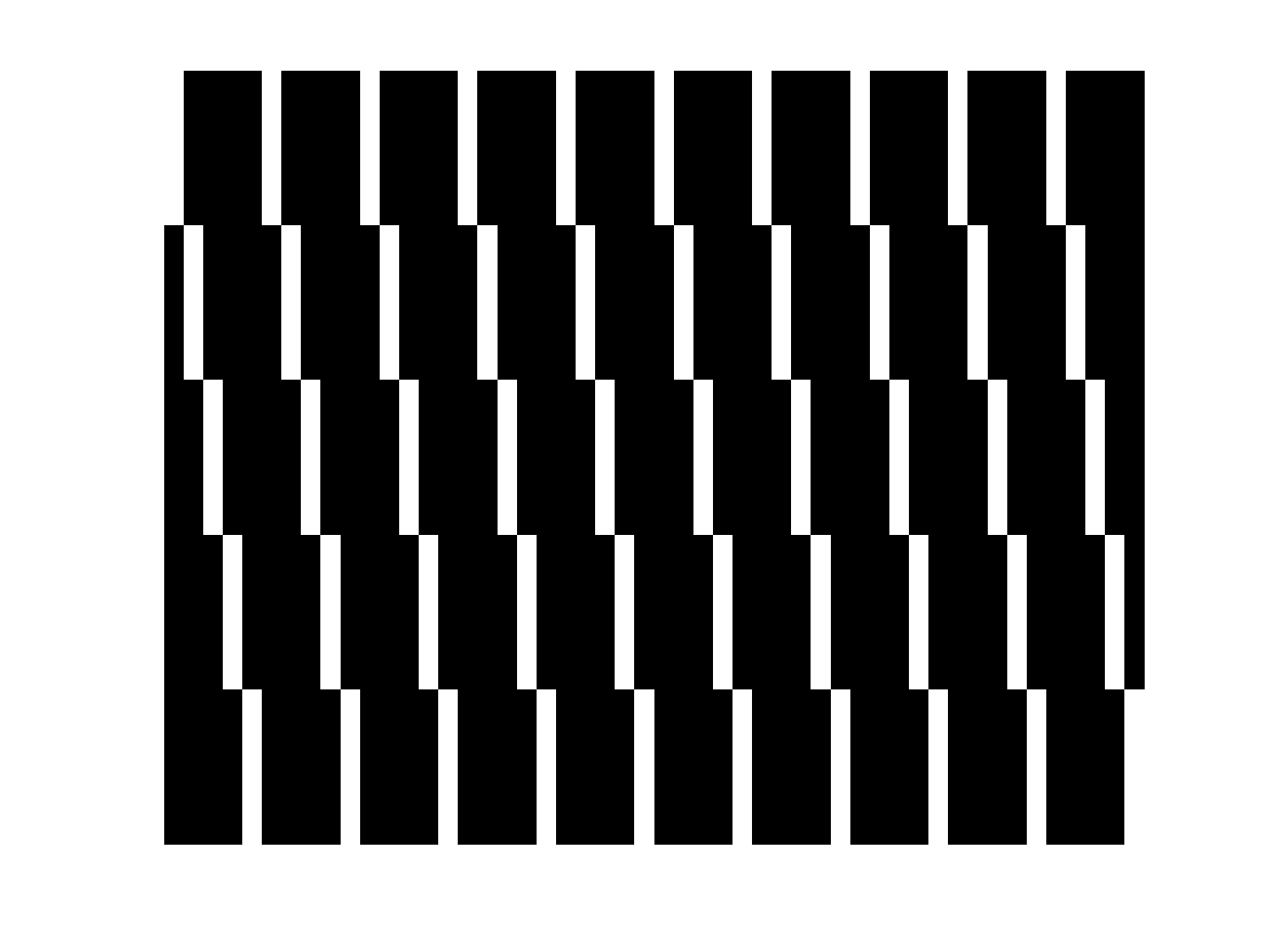}
\includegraphics[width=1.3cm]{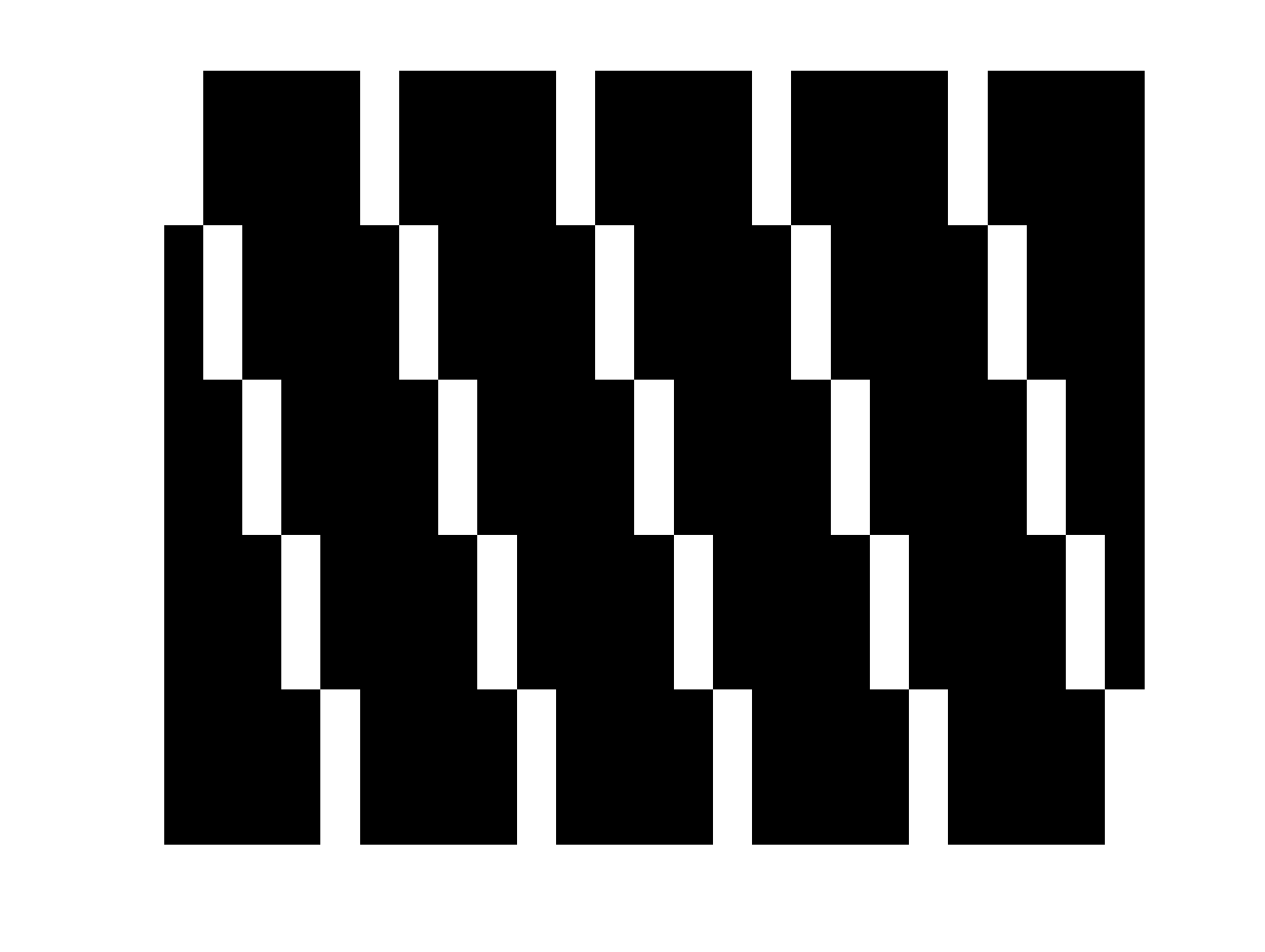}
}
\subfigure[Estimated  $\zeta,\beta,\eta,\tau$]{
\includegraphics[width=1.3cm]{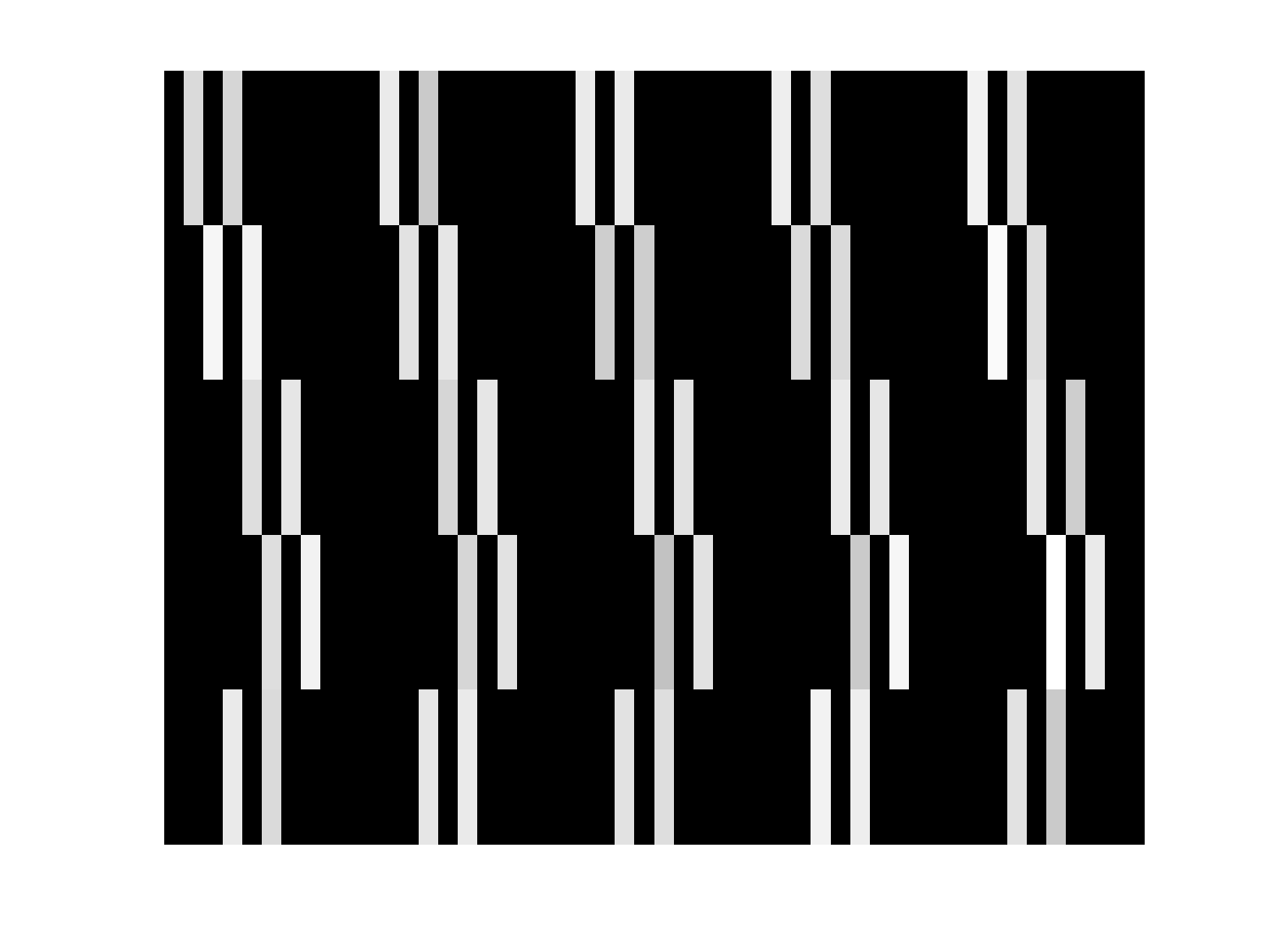}
\includegraphics[width=1.3cm]{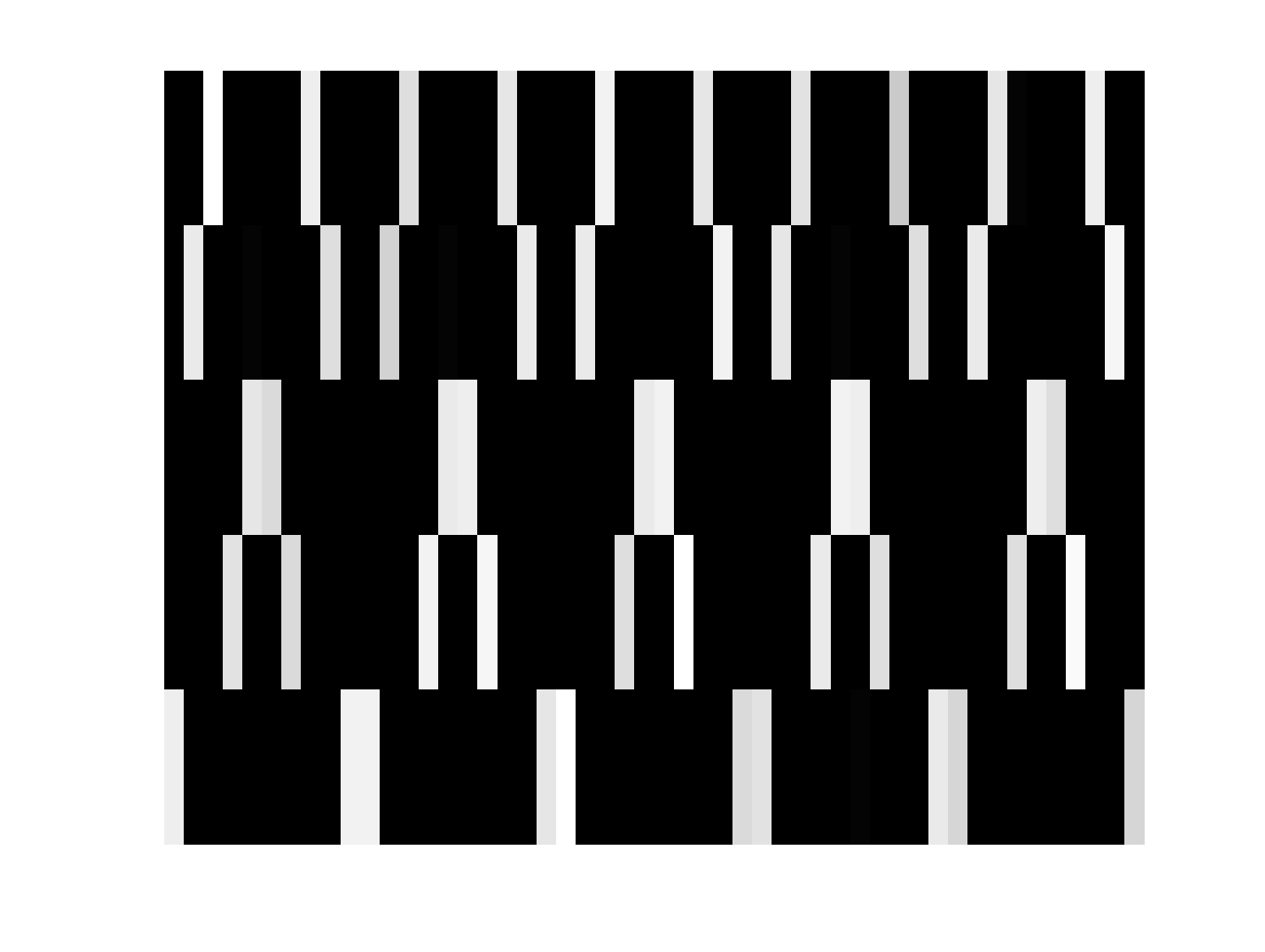}
\includegraphics[width=1.3cm]{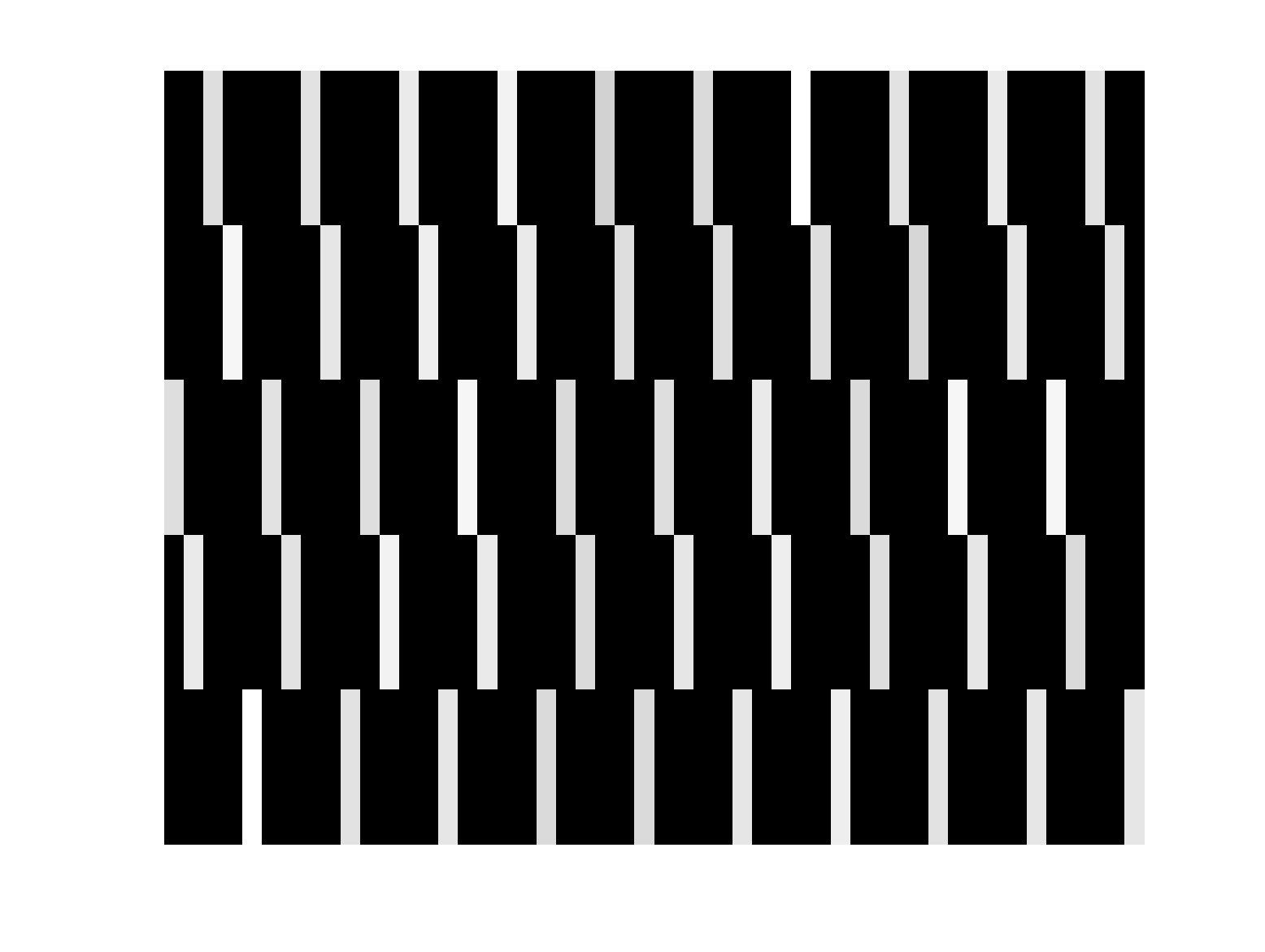}
\includegraphics[width=1.3cm]{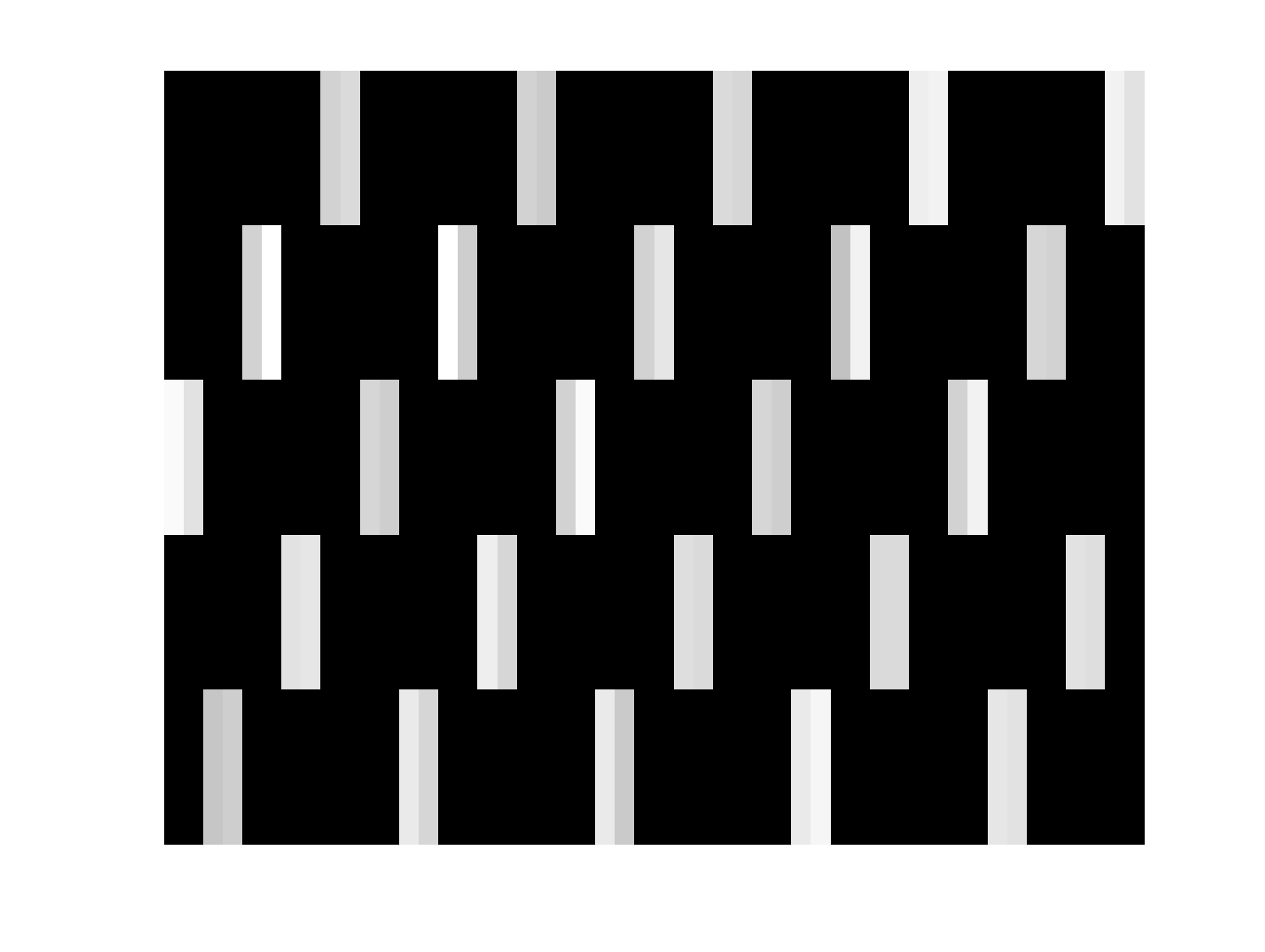}
}
\caption{ \scriptsize  (a) shows the ground truth of the four per topic word distributions in the two-view IBTM model; (b) shows the inferred distributions. Each row in the distributions represents a topic and each column presents a word. There are 5 topics in each distribution ($K=T=S=5$),  and the vocabulary size is 50 for both views ($V=W=50$).  }
\label{fig:synTopics}
\end{minipage}%
~~~~~~
\begin{minipage}{.45\textwidth}
\subfigure[Estimation of $\rho$]{
\includegraphics[width=2.5cm]{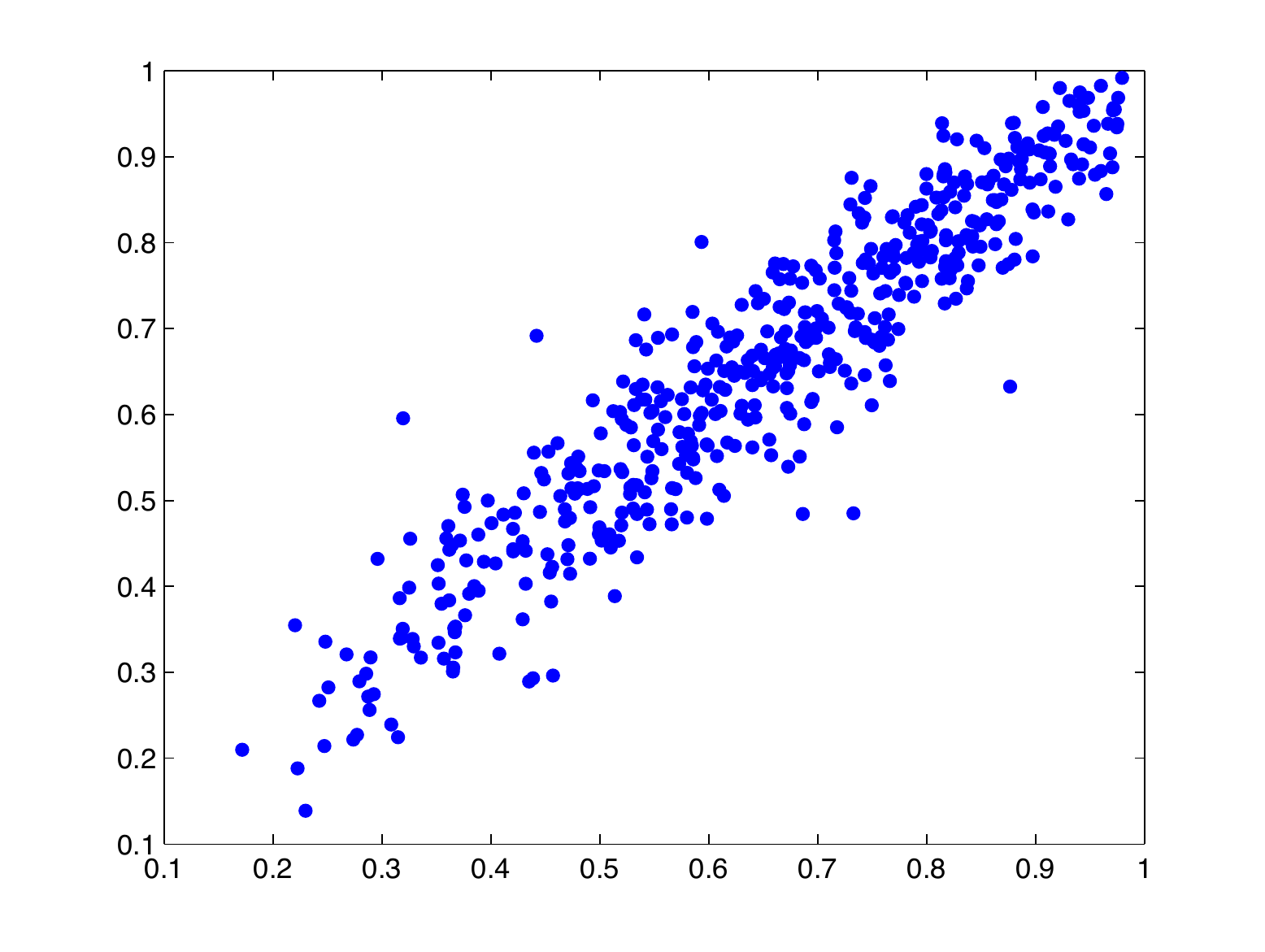}
}
\subfigure[Estimation of $\mu$]{
\includegraphics[width=2.5cm]{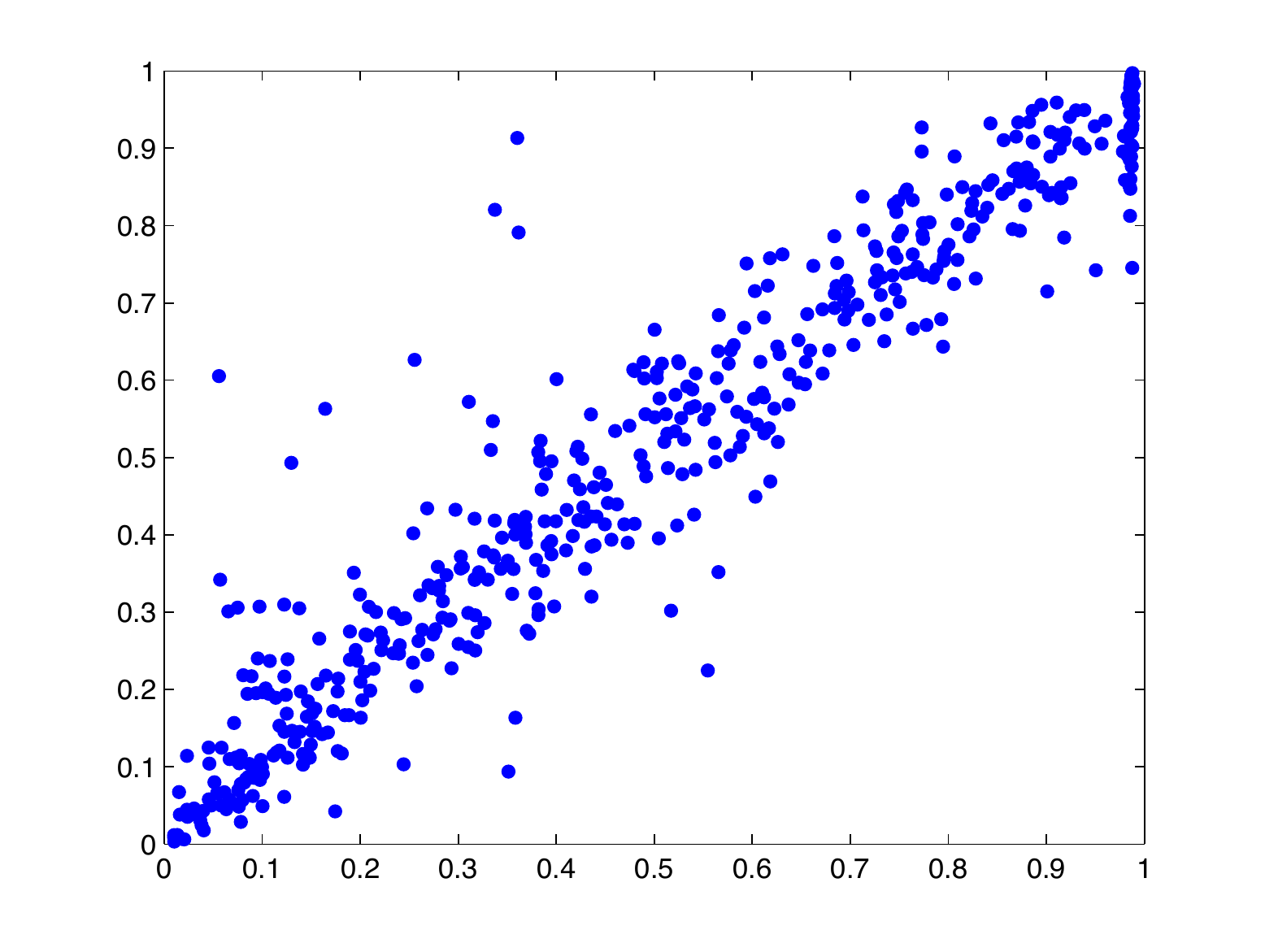}
}
\caption{\scriptsize  Visualization of synthetic data experiment on inference of partition parameters $\rho$ and $\mu$. The x-axis is the ground truth and the y-axis is the estimation. Each dot in the plot presents the $\rho$ and $\mu$ for a document.}
\label{fig:SynBeta}
\end{minipage}
\end{figure}

\subsection{Inference Evaluation using Synthetic Data}
\label{sec:exp_syn}

To test the inference performance, we generate a set of  synthetic data using the model given different topic distributions $\zeta, \beta, \eta, \tau$ and hyper-parameters for $\mu, \rho, \kappa, \theta, \nu$.  We generate 500 documents and each document has 100 words for each view. Given the generated data, a correct inference algorithm will be able to recover all the latent parameters.  Figure \ref{fig:synTopics} (a) shows the ground truth that we used for the per topic words distribution and the estimation of these latent variables using variational inference as described in Section \ref{sec:VB}. All the topics are correctly recovered. Due to the exchangeability of Dirichlet distribution, the estimation gives different order of the topics which is shown as row-wise exchanges in Figure \ref{fig:synTopics}(b).  Figure \ref{fig:SynBeta} shows the parameter recovery for the partition parameters $\rho$ and $\mu$ which are generated from beta distribution. In the example, we use $\iota_1=(4,2)$ and $\iota_2=(1,1)$ as hyper-parameters for the beta distributions. In this setting, the first view is comparably clean; the second view is more noisy with big variations on the noise level among the data. As Figure \ref{fig:SynBeta} shows, almost all the partition parameters are correctly recovered. 

%%%%%%%%%%%%%%%
\subsection{Performance Evaluation using Real-World Data}
\label{sec:exp_bm}

In this section, model performance is evaluated on real-world data. We present two experimental groups. The first one is using the LabelMe natural scene dataset \cite{wang2009simultaneous,li2012objects} and the second one is using the Leeds butterfly dataset \cite{wang2009Learning} for fine-grained classification.  We focus on the model performance where we investigate the distribution of topics and partition parameters. This will provide us with insight into the data structure and model behavior. Thereafter, we will present the classification performance. In these experiments, the classification results are obtained by applying softmax regression on the topic representation. In all experimental settings,  the hyper-parameters for the per document topic distributions are set to $\alpha_{*}=0.8$, the hyper-parameters for the per topic word distributions are set to $\sigma_{*}=0.6$ and the hyper-parameters for the partition variables are set to $\iota_{*}=(5,5)$\footnote{\scriptsize  $\alpha_{*}$ includes $\alpha_{p_1}$,$ \alpha_{s}$ and $\alpha_{p_2}$. $\sigma_{*}$ includes $\sigma_{p_1}$, $\sigma_{p_2}$, $\sigma_{s_1}$ and $\sigma_{s_2}$. $\iota_{*}$ includes  $\iota_1$ and $\iota_2$. }. We also perform experiments with different features, including off-the-shelf CNN-features from different layers and traditional SIFT features. Here, we only present the results using off-the-shelf CNN conv5\_1 features as an example. We use the pre-trained Oxford VGG 16-layer CNN \cite{Simonyan14c} for feature extraction.  We create sliding windows in 3 scales with a 32 pixels step size to extract features, in the same manner as \cite{gong2014multi}, and use K-means clustering to create a codebook and represent each image using a bag-of-visual-words. The vocabulary size is 1024.  In general, the performance is robust when higher layers are used and when the vocabulary size is sufficient. More results using different features and different parameter settings  are enclosed in the supplementary material. 

%%%%%%%%
\subsubsection{LabelMe Dataset.}
We use the LabelMe Dataset as in \cite{wang2009simultaneous,zheng14topic} for this group of the experiments. The LabelMe dataset contains 8 classes of $256 \times 256$ images: highway, inside city, coast, forest, tall buildings, street, open country and mountain. For each class, 200 images are randomly selected, half of which  are used for training, and half of which  are used for testing. This results in 800 training and 800 testing images. We perform the experiment in two different scenarios: Image and Image, where only images are available; and Image and Annotation, where different modalities are available. 
%%%%
\paragraph{Image and Image.}
\label{sec:ImageAndImage}
In this experiment, we explore the scenario in which  only one modality is available. We want to model essential information that captures the within class variations and explains away the instance specific variations. Both views are bag-of-CNN Conv5\_1 feature representations of the image data. For each document, two training images from the same class are randomly paired. This represents the scenario as shown in the introductory Figure \ref{fig:introPic} (a). For the experimental results presented below, the numbers of topics are set to $K=15$, $T=15$, $S=15$.\footnote{\scriptsize  The performance is robust with a sufficient amount of topics, $15$ or higher. More results with different numbers of topics are presented in the supplement.}   

\begin{figure}[p]
\subfigure[$\theta$, Shared]{
\includegraphics[width=3.8cm]{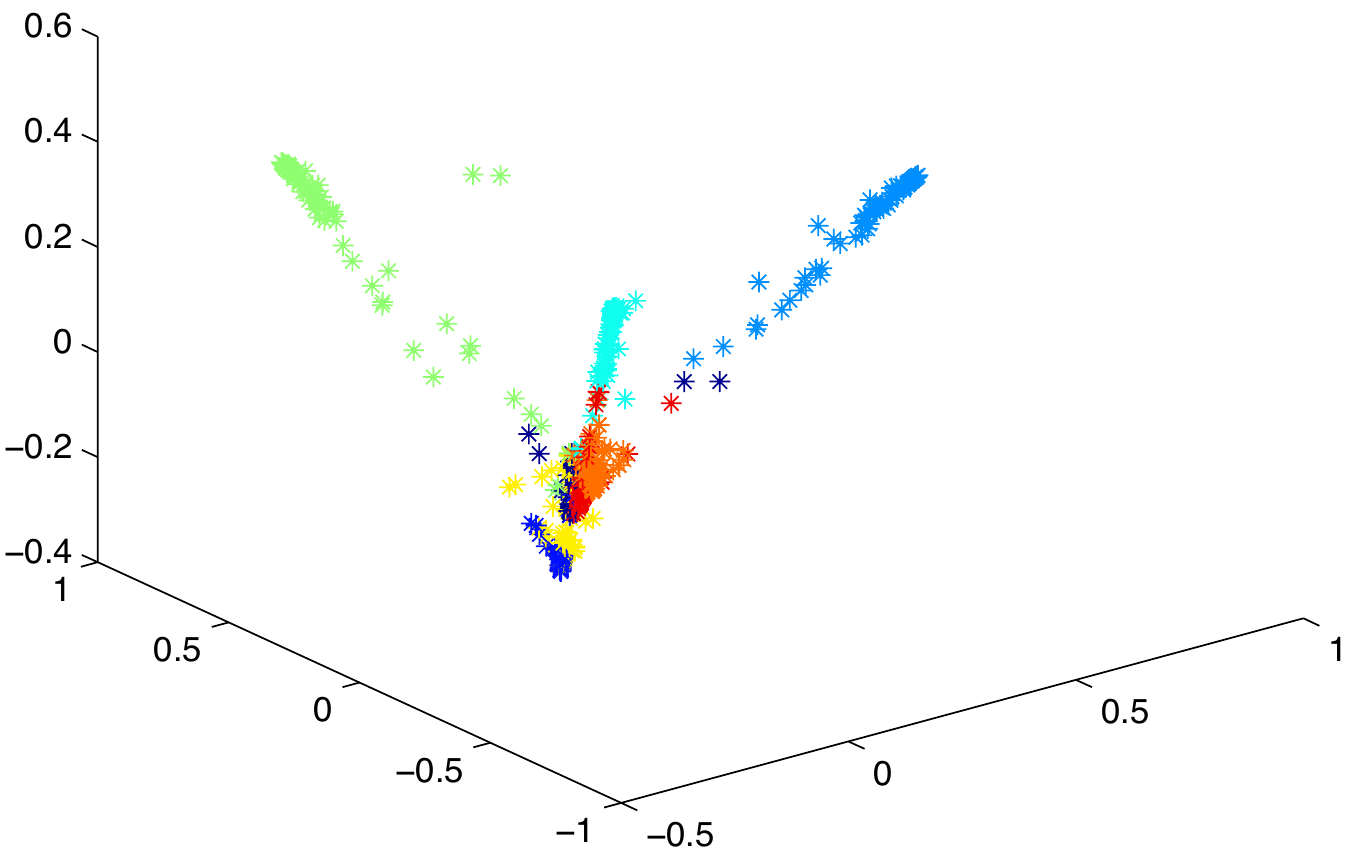}
}
\subfigure[$\kappa$, Private]{
\includegraphics[width=3.8cm]{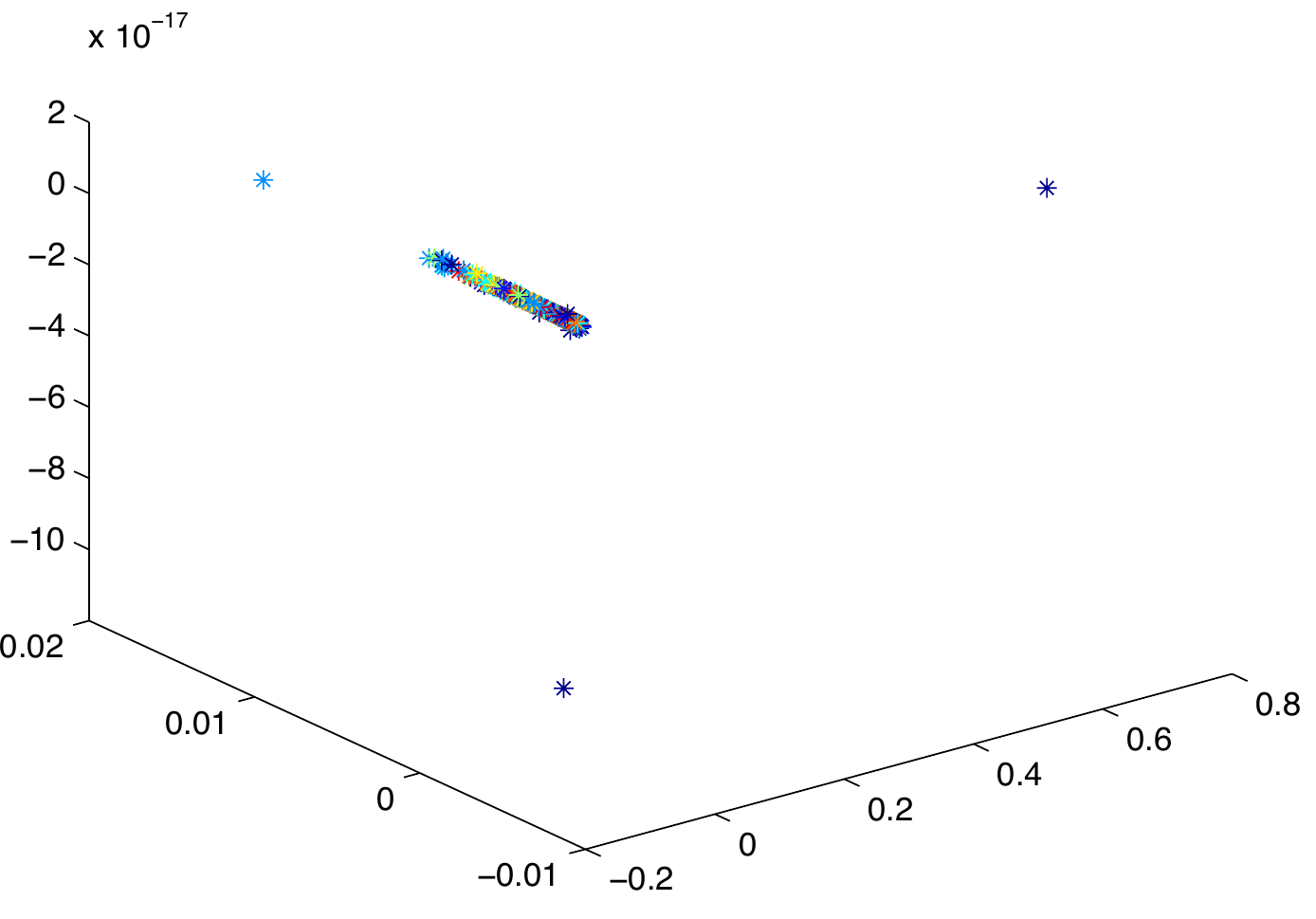}
}
\subfigure[$\nu$, Private]{
\includegraphics[width=3.8cm]{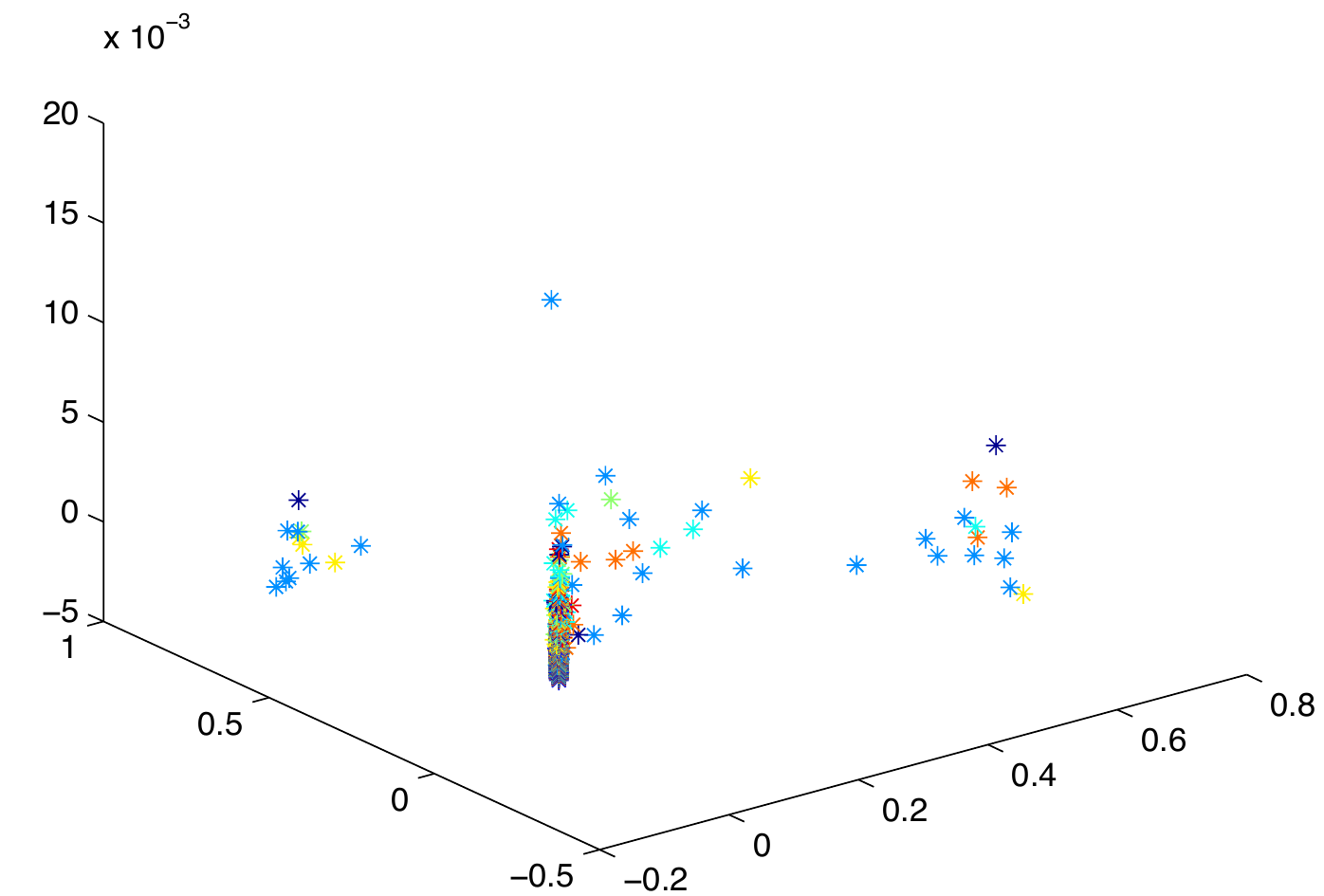}
}
\caption{\scriptsize  Visualization of the shared topic representation ($\theta$) and private topic representations ($\kappa$ and $\nu$) for LabelMe experiments using randomly paired images from the same class. The documents of different classes are colored differently and the plots show the first three principal components after applying PCA on the per document topic distributions for all the training data. }
\label{fig:doc_topic_imageImage_Conv5}
\end{figure}
\begin{figure}[p]
\centering
\begin{minipage}{.48\textwidth}
\subfigure[Hist of $\rho$  (Img) ]{
\includegraphics[width=2.7cm]{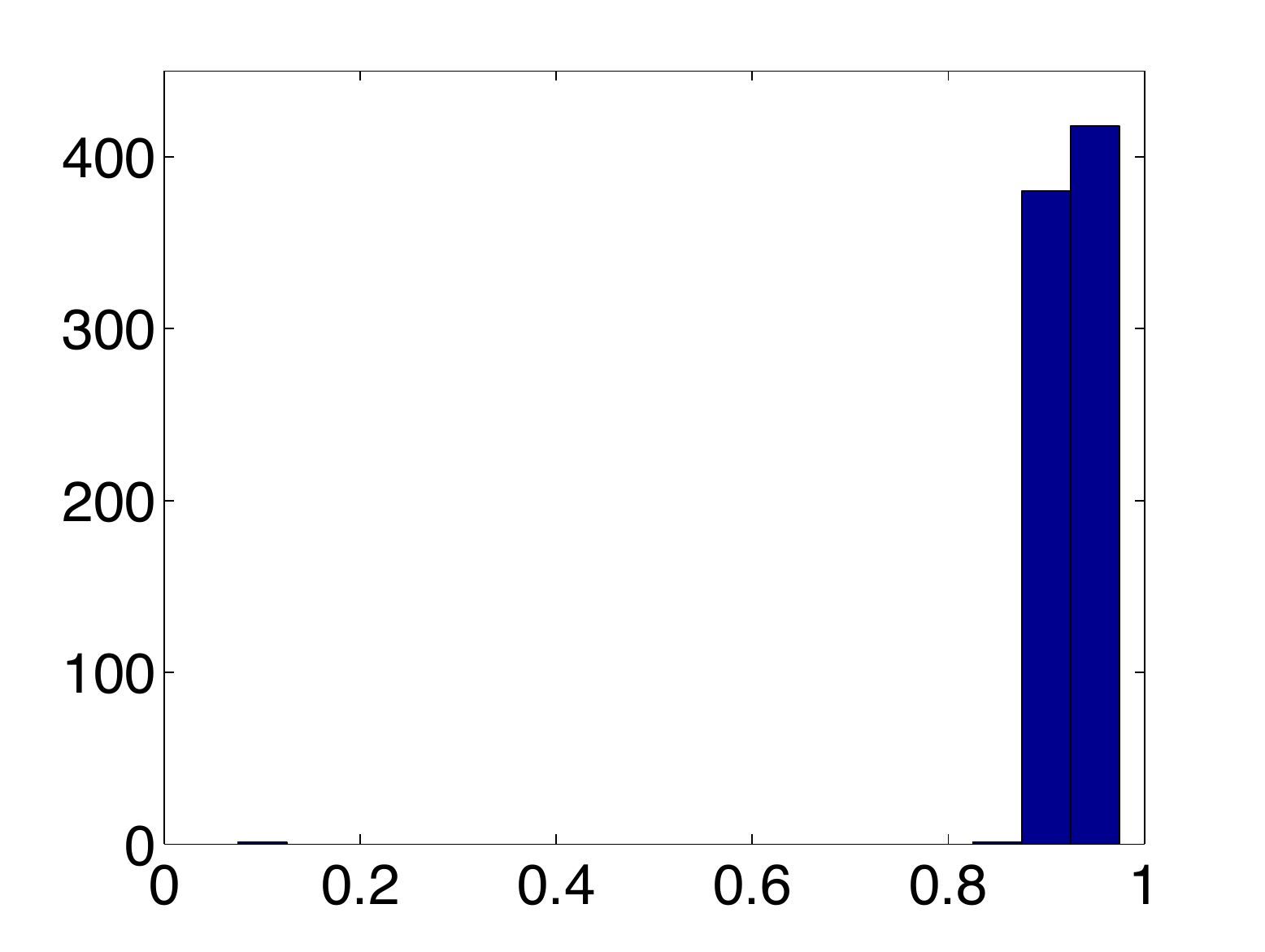}
}
\subfigure[Hist of $\mu$  (Img) ]{
\includegraphics[width=2.7cm]{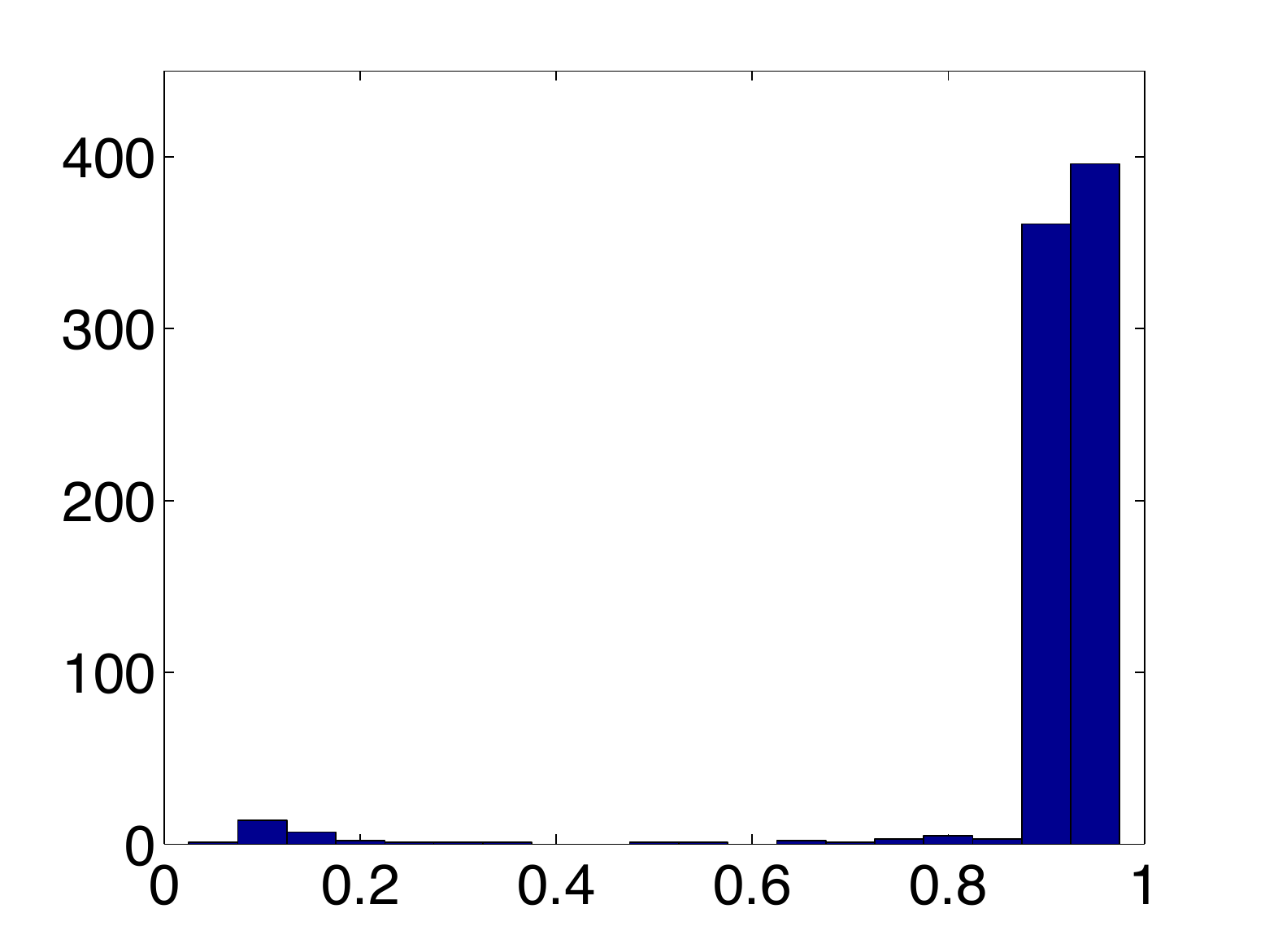}
}
\caption{ \scriptsize  The histogram over partition parameters of the LabelMe image-image experiment. Img indicates that this modality uses natural images.}
\label{fig:hist_imageimage}
\end{minipage}
~~
\begin{minipage}{.48\textwidth}
\subfigure[Hist of $\rho$  (Img) ]{
\includegraphics[width=2.7cm]{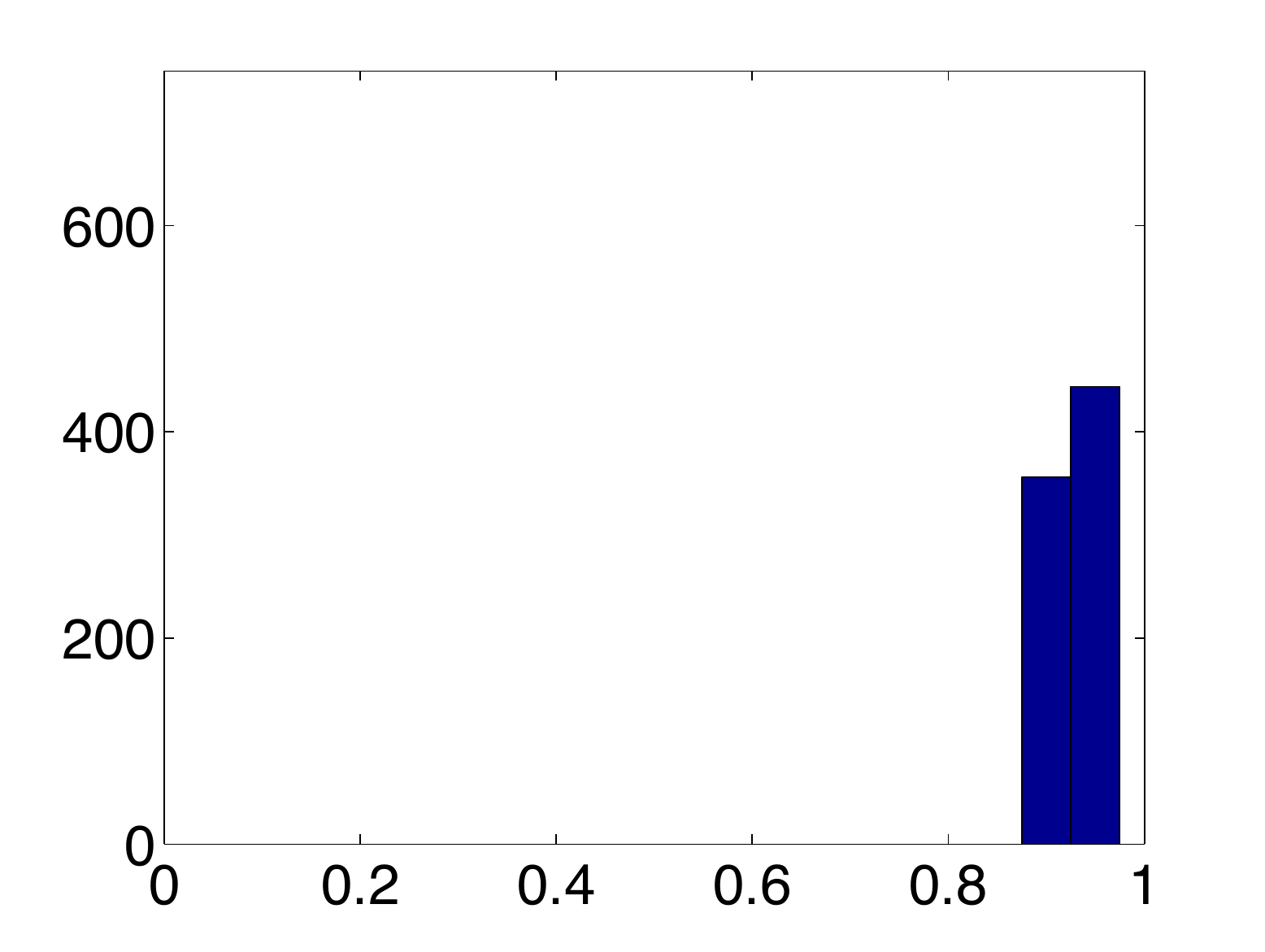}
}
\subfigure[Hist of $\mu$  (Ann) ]{
\includegraphics[width=2.7cm]{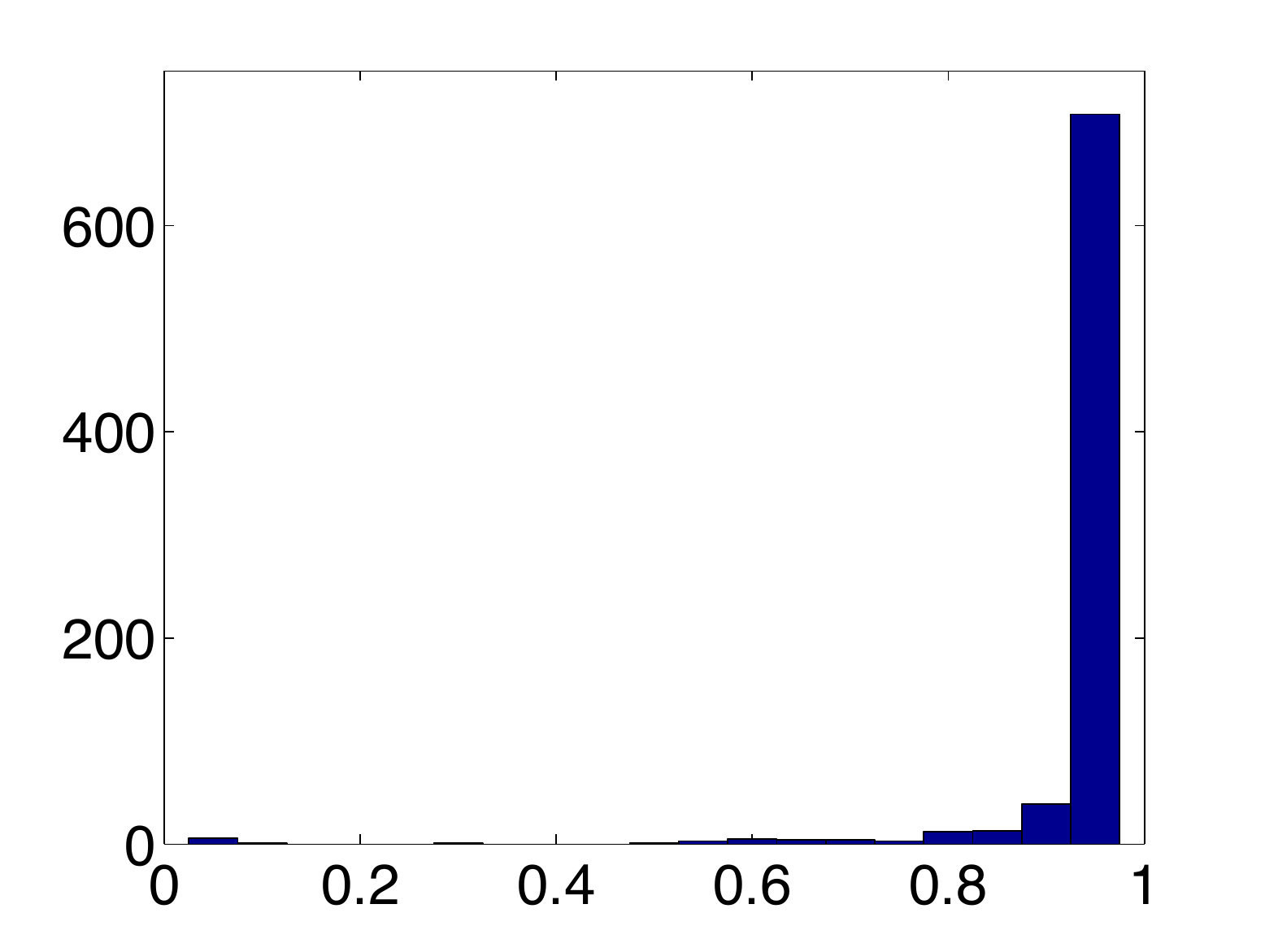}
}
\caption{\scriptsize  The histogram over partition parameters of the LabelMe image-annotation experiment. Img indicates that this modality uses natural images. Ann indicates that this modality uses image annotations.}
\label{fig:hist_imageAnnoLabelMe}
\end{minipage}
\end{figure}

Figure \ref{fig:hist_imageimage} shows the histograms of the partition parameters in this case. Figure \ref{fig:hist_imageimage} (a) and (b) appear to be similar. This is according to intuition; since both views are images and they are randomly paired within the same classes, the statistical features are expected to be the same for both views. Most partition parameters are larger than 0.8, which means that large parts of information can be shared between images from the same class and that the CNN Conv5\_1 features provide a good raw representation of the images. For image pairs with more variation that does not correlate with the image class, the partition parameters will be smaller. The essential information ratio varies among images which causes the partition parameters  to vary among different images. 

\begin{table}[p]
\begin{center}
  \begin{tabular}{ |p{1.9cm}| p{2.0cm} |  p{1.3cm} | p{1.5cm}| p{1.2cm} | p{1.4cm} | p{1.2cm} | }
    \hline
    \scriptsize DocNADE \cite{zheng14topic}&\scriptsize  SupDocNADE\cite{zheng14topic}&\scriptsize Full SVM&\scriptsize  PCA15 SVM&\scriptsize   LDA15 &\scriptsize  SWB15  \cite{chemudugunta2006modeling} & \scriptsize IBTM15  \\
     \hline
     $81.97\%$ & $83.43\%$&$87\%$&$80.88\%$ &$85.25\%$ & $59.88\%$&$\mathbf{89.75\%}$\\
     \hline
  \end{tabular} 
  \end{center}
  \caption{\scriptsize  The performance comparison for Image and Image experiment with the LabelMe dataset.  Full SVM shows the performance using SVM on the bag of Con5\_1 features, while PCA 15 SVM shows the performance after applying PCA and using the top 15 principal components. LDA 15 shows the result using LDA with 15 topics and classification by softmax regression. IBTM 15 shows the result using IBTM with 15 shared topics and classification by softmax regression only on the shared topics. }
  \label{tab:LabelMeImgImg}
\end{table}

Figure \ref{fig:doc_topic_imageImage_Conv5} visualizes the document distribution in different topic representation spaces.  Figure \ref{fig:doc_topic_imageImage_Conv5} (a) shows that documents from different classes are well separated in the space defined by the shared topic representation. Figure \ref{fig:doc_topic_imageImage_Conv5} (b) and (c) show that documents from different classes are more mixed in the private topic spaces. Thus, the private information is used to explain instance specific features of a data point, but not  class-specific features -- these have been pushed into the shared space, according to the intention of the model. The variations in the private spaces are small due to the low noise ratio in the dataset. For the classification performance where only images are available, using IBTM with classification using only the shared representation leads to a classification rate of $89.75\%$. The classification results are summarized in Table \ref{tab:LabelMeImgImg}. A standard LDA obtains better performance than PCA with the same number of dimensions. IBTM outperforms LDA with the same number of topics and can even obtain better results than using the full dimension (1024) of bag-of-Conv5\_1 features together with linear SVM. While using SWB \cite{chemudugunta2006modeling} \footnote{ \scriptsize  We implemented SWB using Gibbs Sampling following the description in the paper \cite{chemudugunta2006modeling}. The parameter settings are the same as in \cite{chemudugunta2006modeling}. Linear SVM is used for classification using the topic representation from SWB. More analysis using SWB is presented in the supplementary material of this paper. }, the performance is unsatisfactory for such computer vision tasks due to the noisy properties of images. The results show that IBTM is able to learn a factorized latent representation, which separates task-relevant variation in the data from variation that is less relevant for the task at hand, here classification.

%%%%%%%%%%%%%%%%%
\begin{figure}[p]
\subfigure[$\theta$, Shared]{
\includegraphics[width=3.8cm]{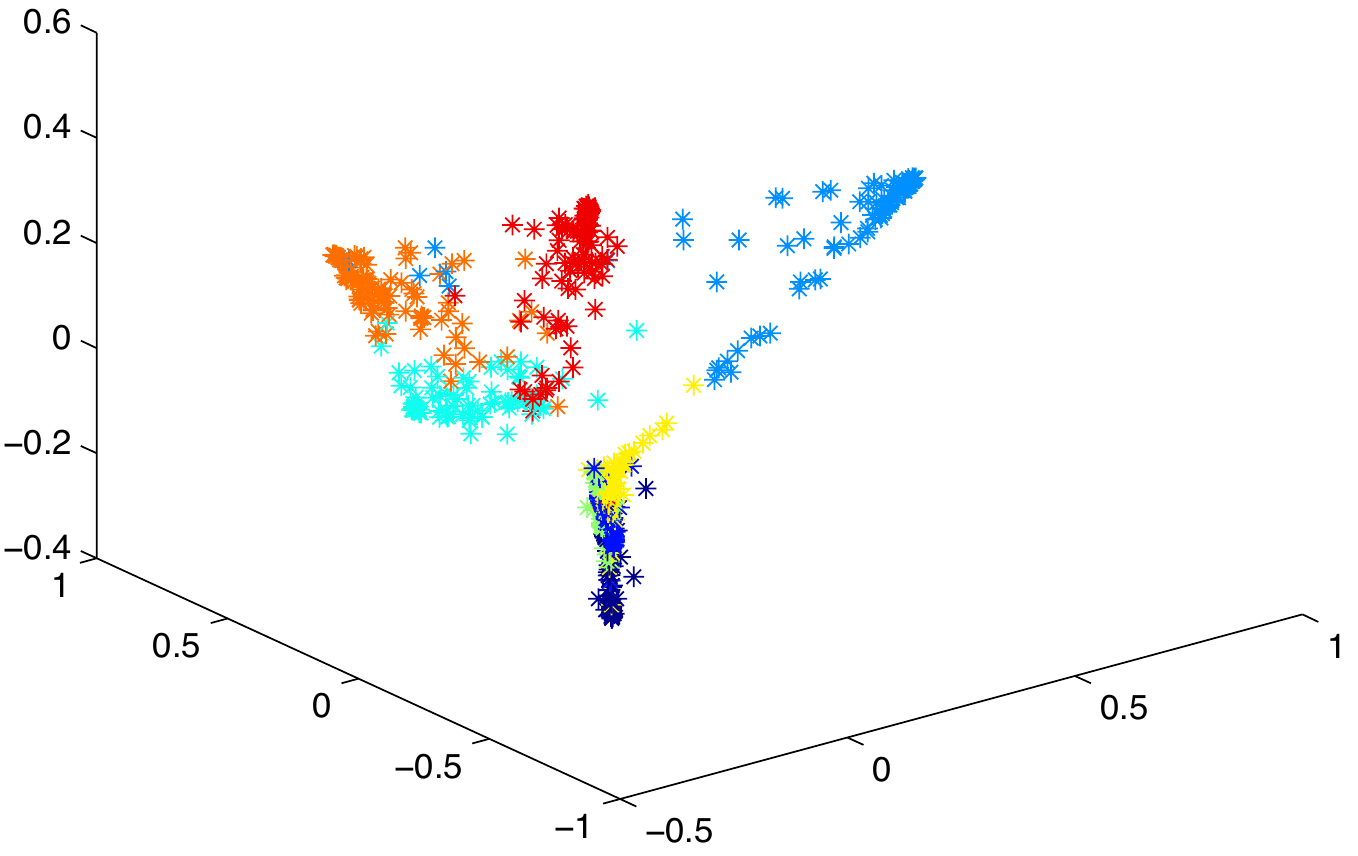}
}
\subfigure[$\kappa$, Private]{
\includegraphics[width=3.8cm]{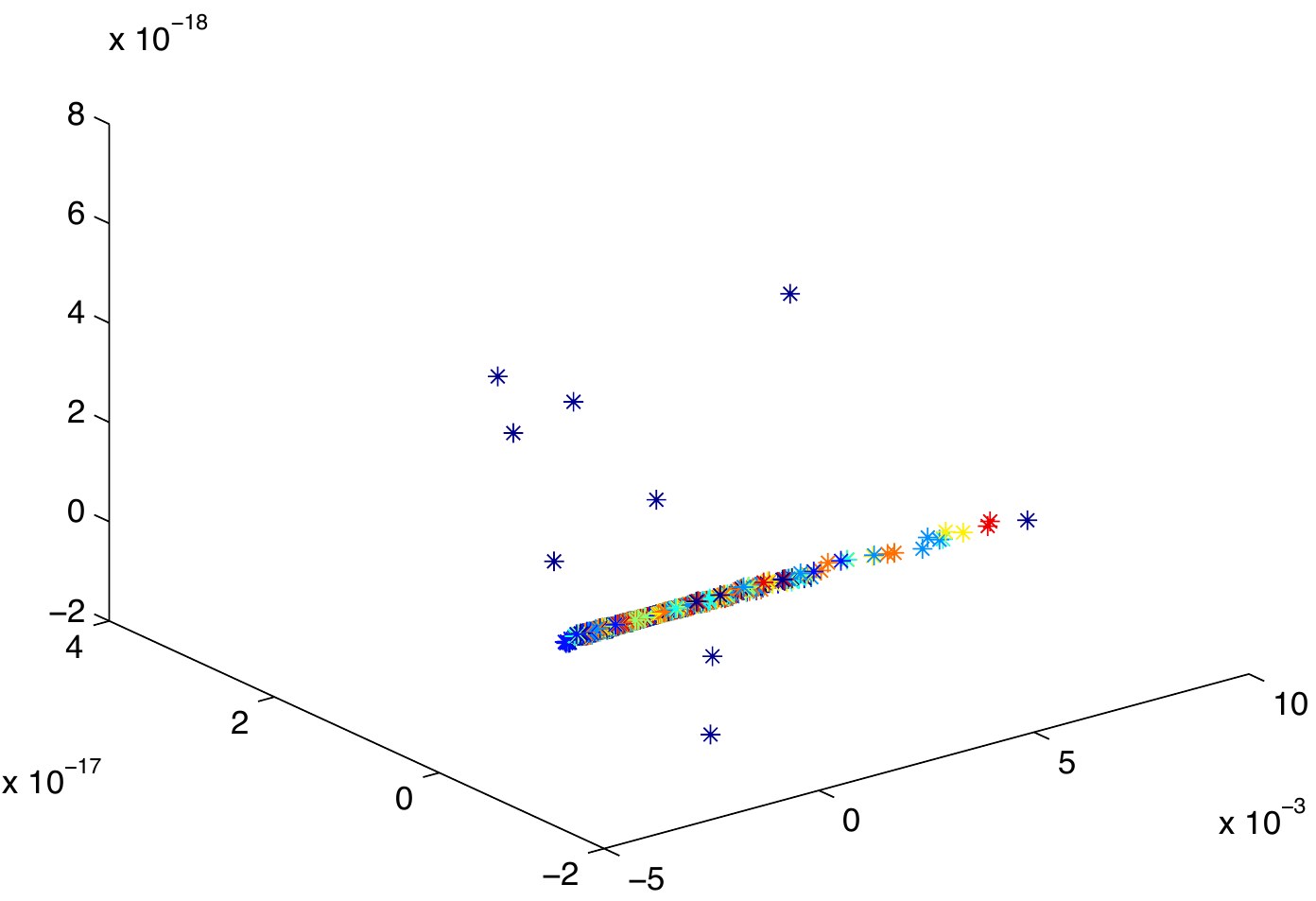}
}
\subfigure[$\nu$, Private]{
\includegraphics[width=3.8cm]{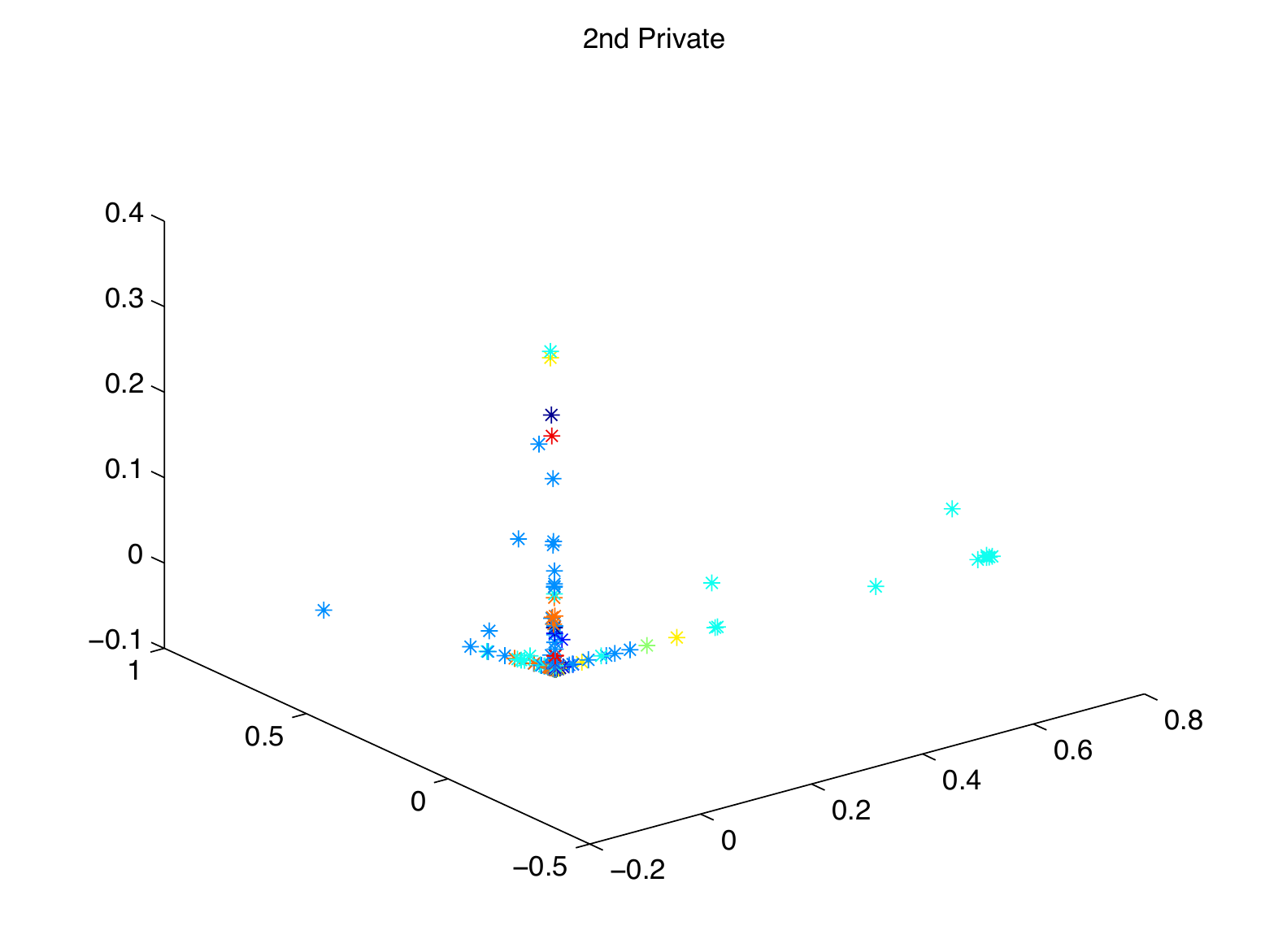}
}
\caption{\scriptsize  Visualization of the shared topic representation ($\theta$) and private topic representations ($\kappa$ and $\nu$) for LabelMe experiments using image features for the 1st view and annotation for the 2nd view.The documents of different classes are colored differently and the plots show the first three principal components after applying PCA on the per document topic distributions for all the training data. }
\label{fig:doc_topic_imageAnno_Conv5LabelMe}
\end{figure}

\paragraph{Image and Annotation.}
\label{sec:ImageAndAnnot}
In this experiment, we explore the scenario when two different modalities are available for different views. We use the bag-of-Conv5\_1 representation of images as the first view and the image annotations as the second view. The word counts for the annotations are scaled with the annotated region. For each document, 79 Conv5\_1 features are extracted from the image view, and the sum over the word histogram for each view is normalized to 100.  The number of topics is set to $K=15$, $T=15$, $S=15$ in the experimental results presented here.  Figure \ref{fig:hist_imageAnnoLabelMe} shows the histograms of the partition parameters $\rho$ and $\mu$ for the two views respectively. Figure \ref{fig:hist_imageAnnoLabelMe} (b) shows that the partition parameters are more concentrated around large values compared to Figure \ref{fig:hist_imageAnnoLabelMe} (a), which indicate that most annotation information is more essential. This is consistent with the intuition of the relative noise levels in image vs annotation data. 

\begin{table}[ht]
\begin{center}
  \begin{tabular}{ |p{1.8cm}| p{1.2cm} | p{1.2cm} |p{1.8cm} | p{1.8cm} |p{1.8cm}| }
    \hline
    Full SVM  & PCA 15& LDA15 & SWB 2V  \cite{chemudugunta2006modeling} &IBTM15  1V &IBTM15  2V  \\  
     \hline
     $87.63\%$  & $84.88\%$& $85.38\%$ &$61\%$&$\mathbf{89.38\%}$ &$\mathbf{95\%}$\\
     \hline
  \end{tabular}
  \end{center}
  \caption{\scriptsize  The performance comparison for the image-annotation experiment for the LabelMe dataset. "IBTM15 1V"  shows the prediction performance with only images available (1 view testing)  and "IBTM15 2V"  shows the prediction performance with both  images and annotation  available  (2 view testing).  For "SWB 2V", we concatenate words from images and captions for each document for both training and testing to use SWB since it is a single-view model }
  \label{tab:ImgAnnoLabelMe}
\end{table}

Figure \ref{fig:doc_topic_imageAnno_Conv5LabelMe} shows the distribution of documents using different topic representations. As in the previous experiment, documents from different classes are well separated in the shared topic representation and are more mixed in the private topic representations. Table \ref{tab:ImgAnnoLabelMe} summarizes the classification performance.\footnote{ \scriptsize The $0.65\%$ difference of Full SVM performance in Table \ref{tab:LabelMeImgImg} and Table \ref{tab:ImgAnnoLabelMe}  were due to different random data partitions.} IBTM is able to outperform other methods with a performance of $89.38\%$ even when only images are available for testing. When both modalities are available, the performance goes up to $95\%$, while ideal classification by humans for this dataset is reported to be $90\%$ in \cite{li2012objects}. 
%%%%%%%%

\subsubsection{Leeds Butterfly Dataset.}
In this section, the Leeds butterfly dataset \cite{wang2009Learning} is used to evaluate the IBTM model on a fine-grained classification task. This dataset contains 10 classes of butterfly images collected from Google image search, both the original images with cluttered background and segmentation masks for the butterflies are provided in the dataset. For each class, 55 to 100 images have been collected and there are 832 images in total. In this experiment, 30 images are randomly selected from each class for training and the remaining 532 images are used for testing.  Similarly to above, we perform the experiment in two different scenarios: Image and Image, where only the natural images with cluttered backgrounds are available; and Image and Segmentation, where one modality is the natural image and the other modality is the segmented image.

%%%%

\begin{figure}[ht]
\subfigure[$\theta$, Shared]{
\includegraphics[width=3.8cm]{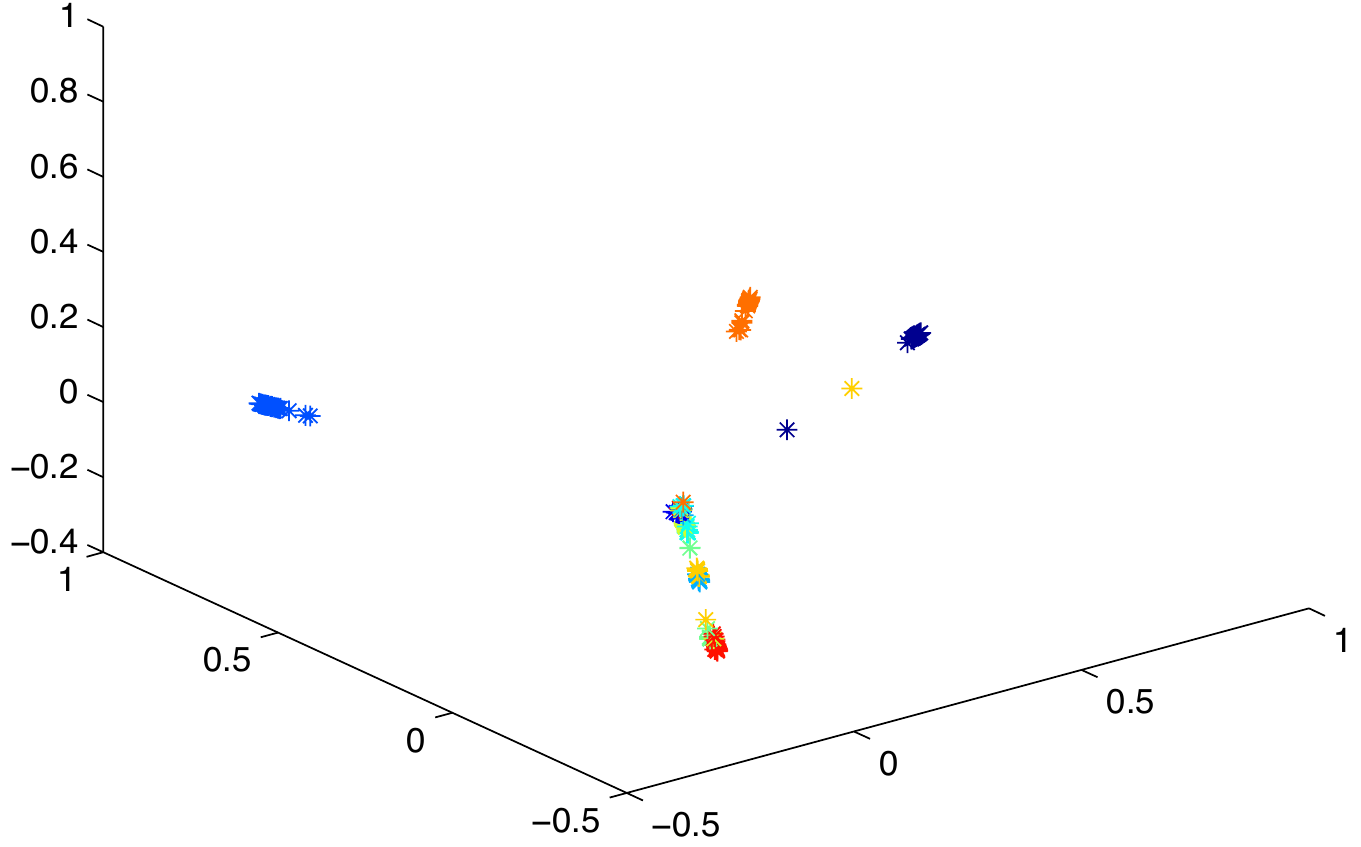}
}
\subfigure[$\kappa$, Private]{
\includegraphics[width=3.8cm]{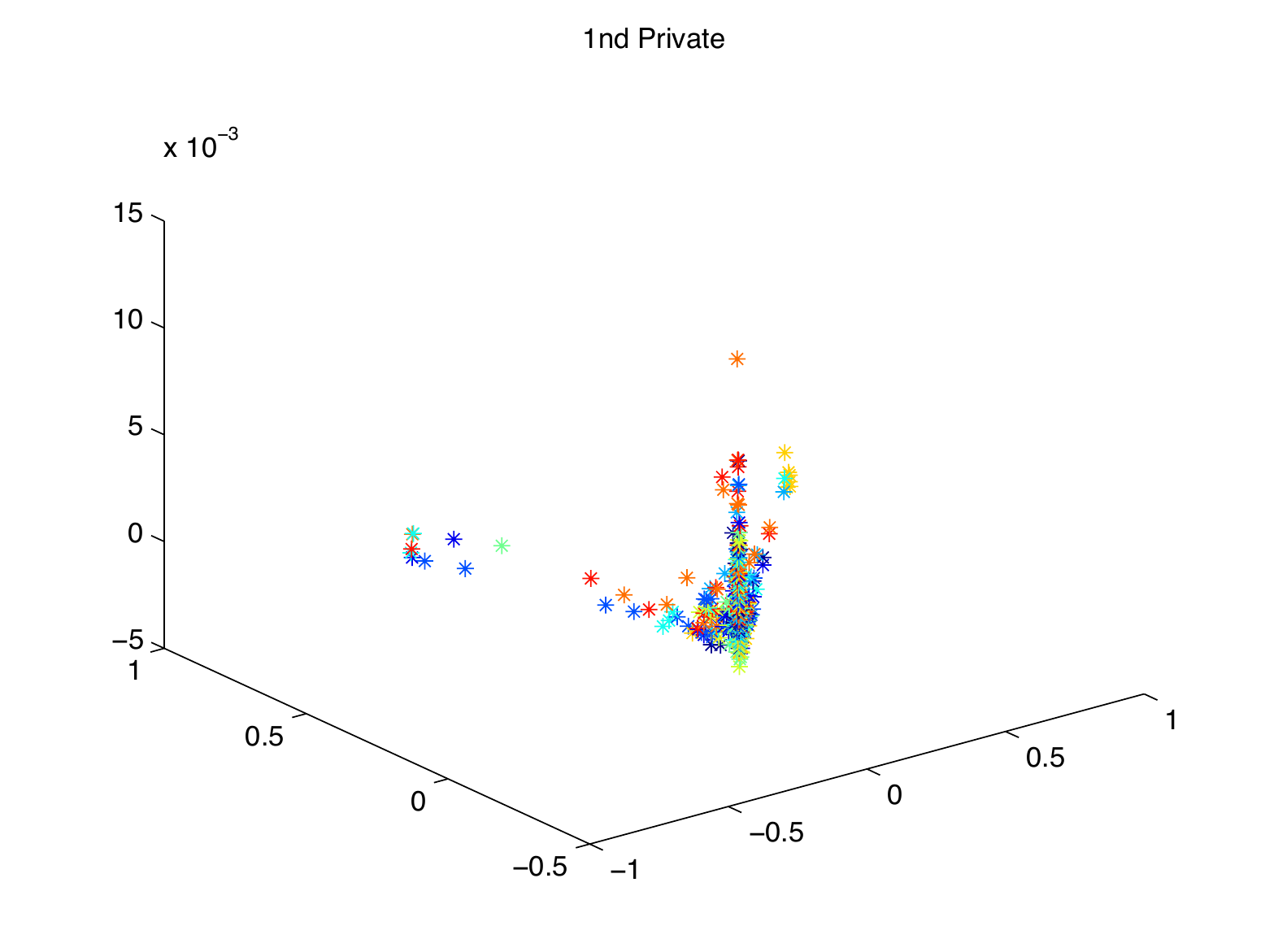}
}
\subfigure[$\nu$, Private]{
\includegraphics[width=3.8cm]{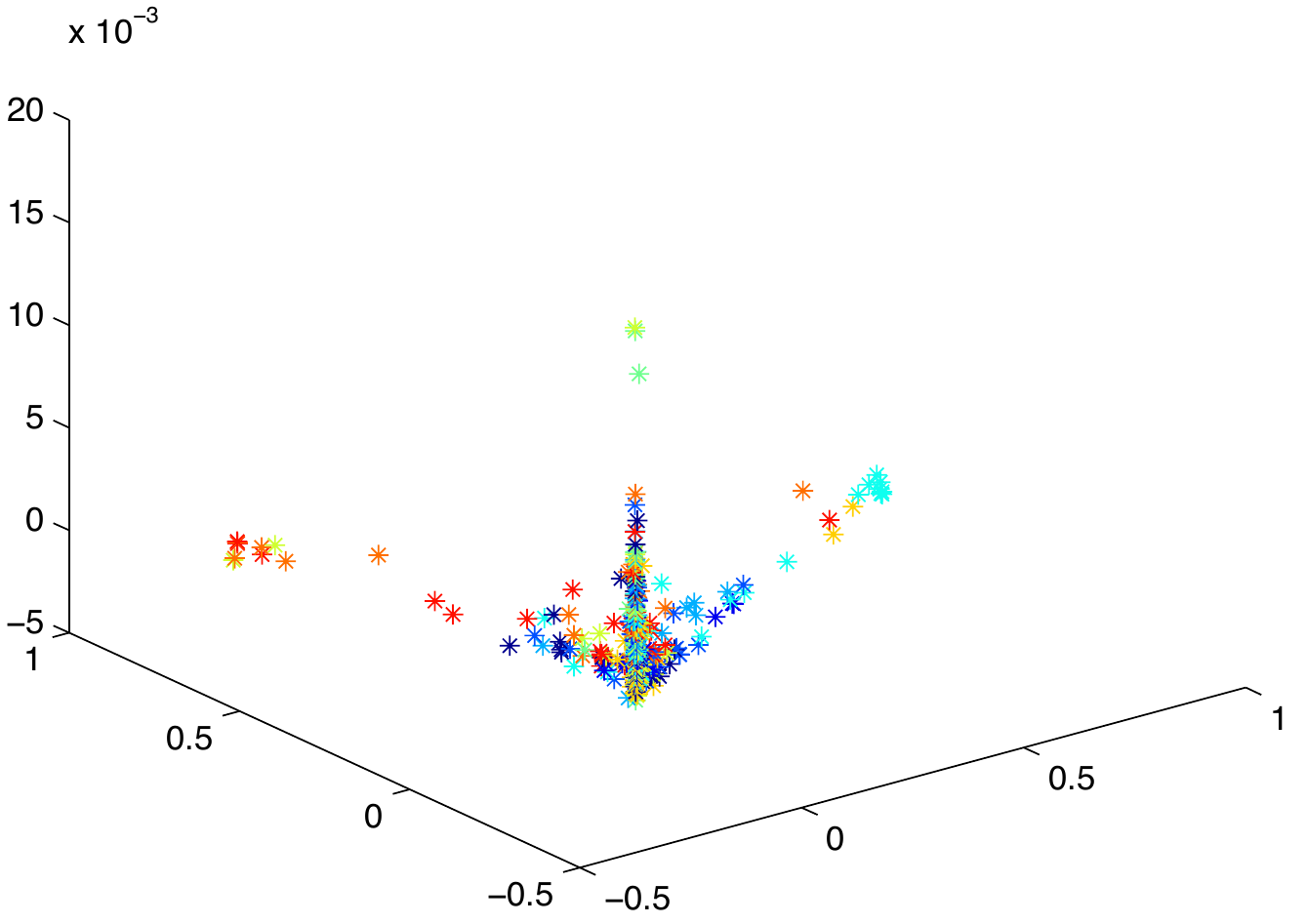}
}
\caption{\scriptsize Visualization of the shared topic representation ($\theta$) and private topic representations ($\kappa$ and $\nu$) for experiments on the Leeds Butterfly dataset using  randomly paired images from the same class. The documents of different classes are colored differently and the plots show the first three principal components after applying PCA on the per document topic distributions for all the training data.  }
\label{fig:doc_topic_ImgImg_Leeds}
\end{figure}

\begin{figure}[ht]
\centering
\begin{minipage}{.48\textwidth}
\subfigure[Hist of $\rho$ (Img) ]{
\includegraphics[width=2.7cm]{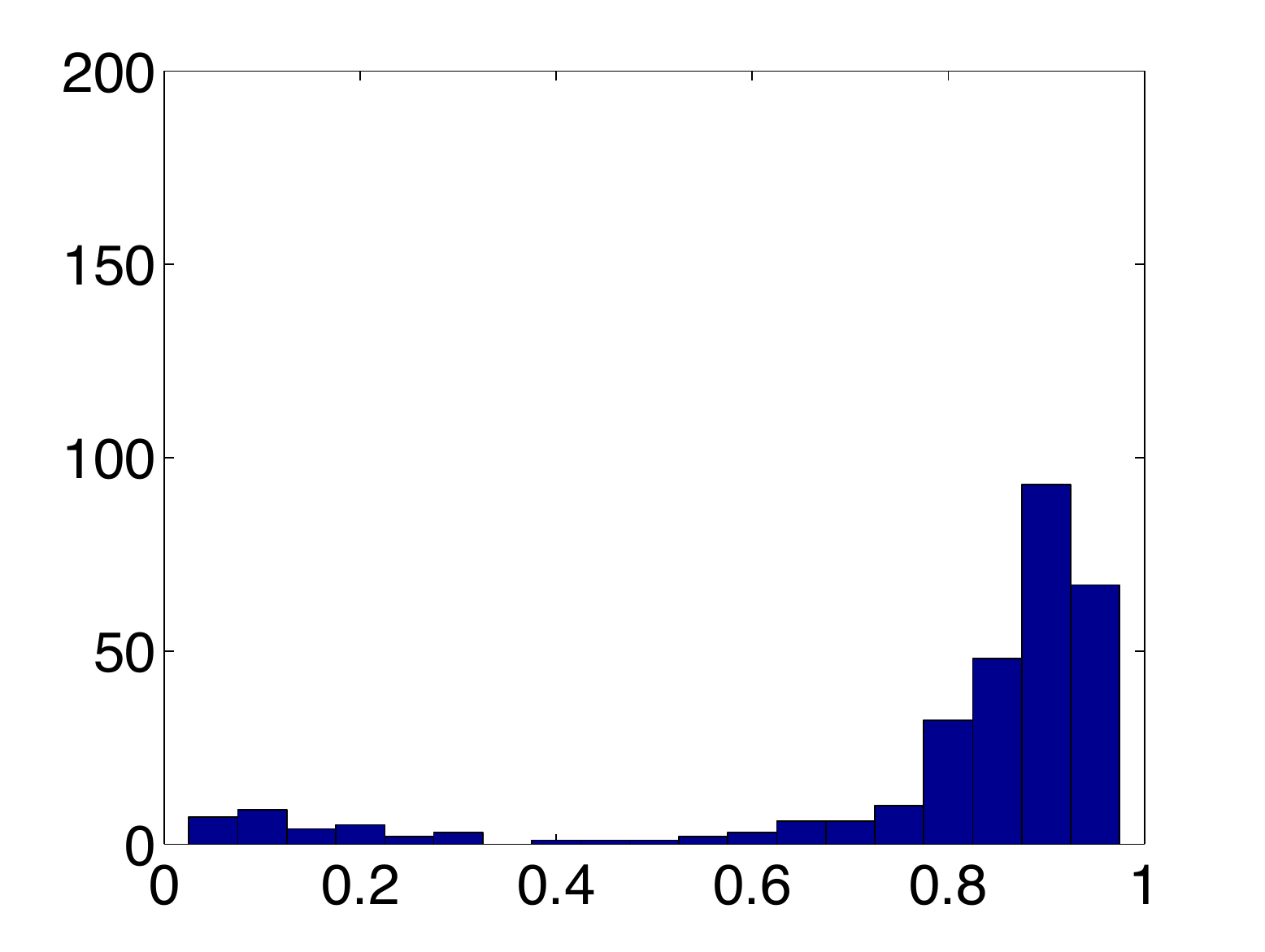}
}
\subfigure[Hist of $\mu$ (Img)  ]{
\includegraphics[width=2.7cm]{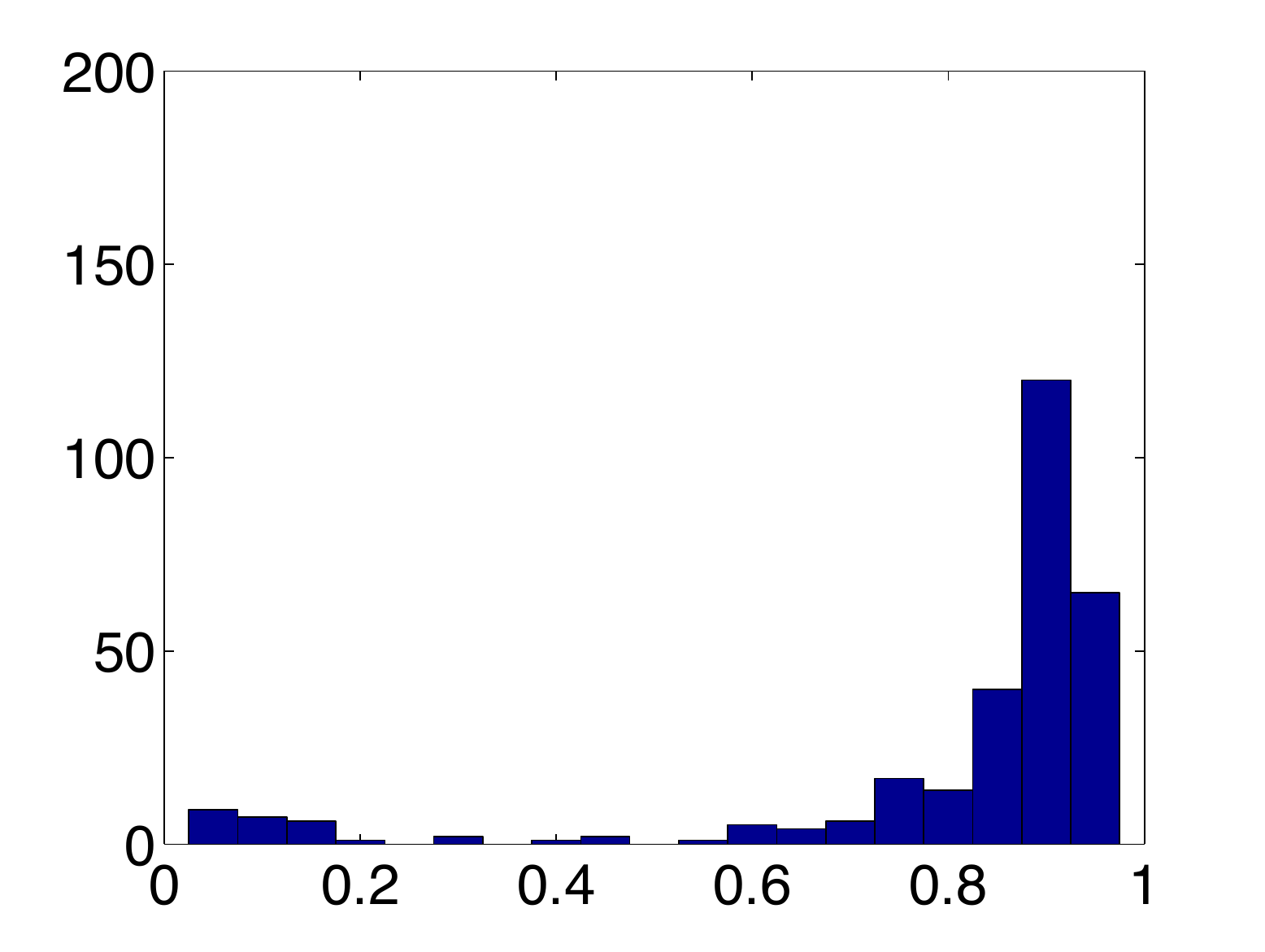}
}
\caption{\scriptsize  The histogram over partition parameters of the Leeds Butterfly image-image experiment. Img indicates that this modality uses natural images.  }
\label{fig:hist_imageImageLeeds}
\end{minipage}
~~
\begin{minipage}{.48\textwidth}
\subfigure[Hist of $\rho$ (Seg) ]{
\includegraphics[width=2.7cm]{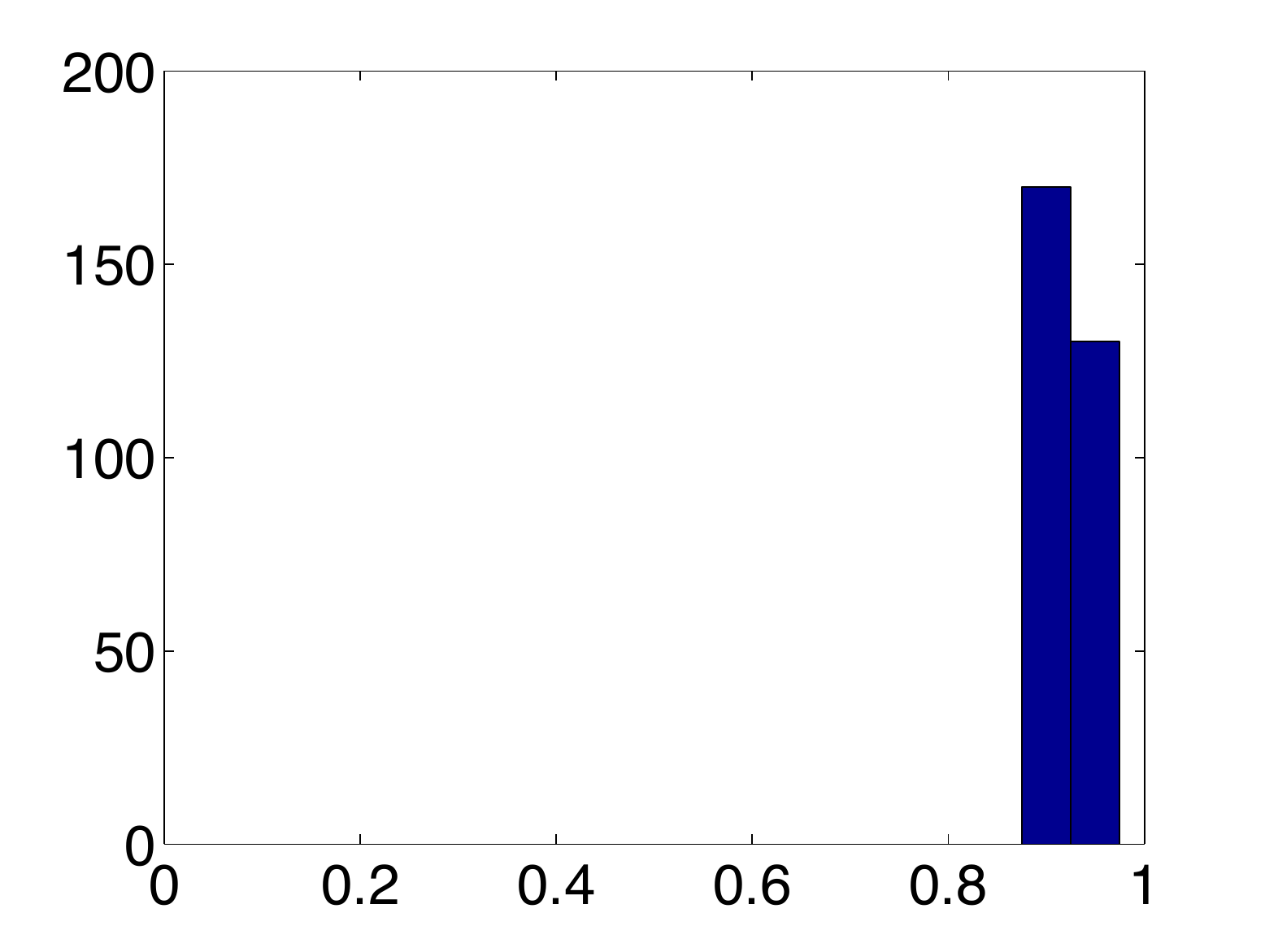}
}
\subfigure[Hist of $\mu$ (Img) ]{
\includegraphics[width=2.7cm]{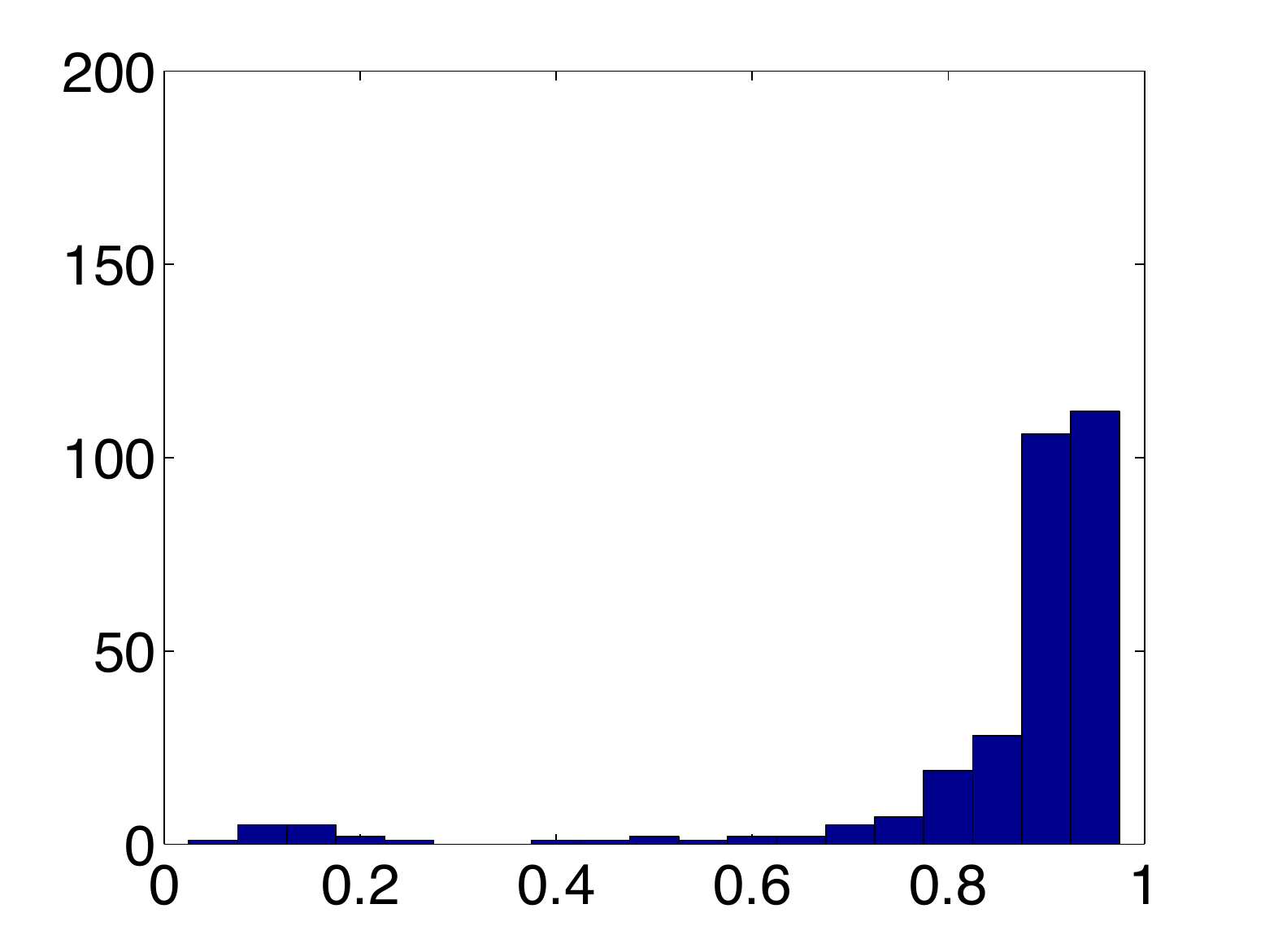}
}
\caption{\scriptsize  The histogram over partition parameters of the Leeds Butterfly image-segmentation experiment. Img indicates that this modality uses natural images. Seg indicates that this modality uses segmented images}
\label{fig:hist_imageSegLeeds}
\end{minipage}
\end{figure}

\paragraph{Image and Image.}
\label{sec:ImageAndImageLeeds}

In this experiment, we use only the natural images to evaluate the model performance in the uni-modal scenario. The experimental setting is similar to Section \ref{sec:ImageAndImage}, where two images from the same class are paired randomly. $K=15$, $T=3$ and $S=3$ are used for the results presented here. The histograms in Figure \ref{fig:hist_imageImageLeeds} are to the previous dataset, however, with smaller values. As natural images of butterflies have more background information that is not related to the class of the butterfly, while for the LabelMe dataset, almost the whole image has information contributing to the natural scene class.

 Figure \ref{fig:doc_topic_ImgImg_Leeds} visualizes the image distribution in the different topic representations, where the shared topic representation separates images from different classes better than the private ones.  Table \ref{tab:leedsdataset} summarizes the classification performance for this dataset. There "II IBTM 15" shows the result of IBTM using only natural images, which obtains the highest performance $95.86\%$ in this uni-modality setting with only 15 topics.

\begin{savenotes}
\begin{table}[ht]
\begin{center}
  \begin{tabular}{ |p{1.1cm}|p{1.1cm}| p{1.05cm} | p{1.45cm}| p{1.53cm} | p{1.05cm} | p{1.2cm} |p{1.3cm} |p{1.3cm}| }
    \hline
      \scriptsize NLD\cite{wang2009Learning} \footnote{ \scriptsize Learning Models for Object Recognition
from Natural Language Descriptions (NLD) trained a classification model based on text descriptors. All images are tested  to use visual information to extract attributes to fit the text template for testing. The experiment setting is different from our experiments. However, we include the result from the original paper for completeness.}&   \scriptsize  Full SVM  &  \scriptsize  PCA 15 &   \scriptsize II SWB15   \cite{chemudugunta2006modeling} &   \scriptsize IS SWB15  \cite{chemudugunta2006modeling}&    \scriptsize LDA15 &   \scriptsize II IBTM15  &   \scriptsize IS IBTM 1V &   \scriptsize IS IBTM 2V \\  
     \hline
     $56.3\%$&$95.49\%$  & $88.35\%$&$80.26\%$ &$94.55\%$ & $91.92\%$&$\mathbf{ 95.86\%}$&$ \mathbf{96.05\%}$&$\mathbf{99.06\%}$\\
     \hline
  \end{tabular}
  \end{center}
    \caption{ \scriptsize The performance comparison with the Leeds Butterfly dataset. "II"  shows the prediction performance  for the paired image setting (Image-Image) for IBTM and only images for SWB.  "IS"  shows the prediction performance  for the  image and its segmentation image setting. In this setting,  "1V" means that  only images are available (1 view testing) and "2V"  means that both  images and segmentations are  available  (2 view testing). "IS SWB" shows the performance of using SWB with concatenated words from images and segmentations. }
  \label{tab:leedsdataset}
\end{table}
\end{savenotes}

%%%%%%%%%%%%%%%
\begin{figure}[h]
\subfigure[$\theta$, Shared]{
\includegraphics[width=3.8cm]{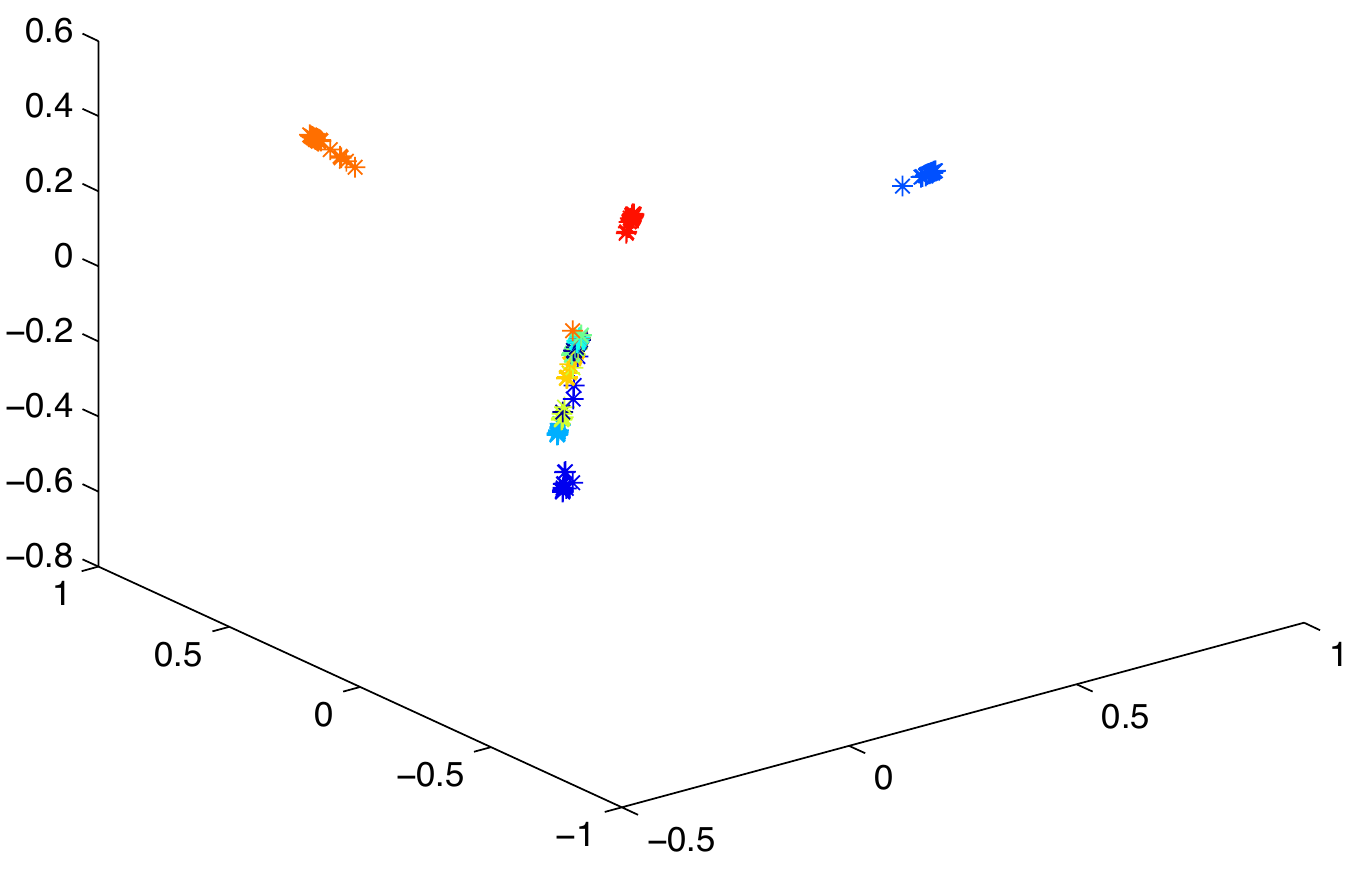}
}
\subfigure[$\kappa$, Private]{
\includegraphics[width=3.8cm]{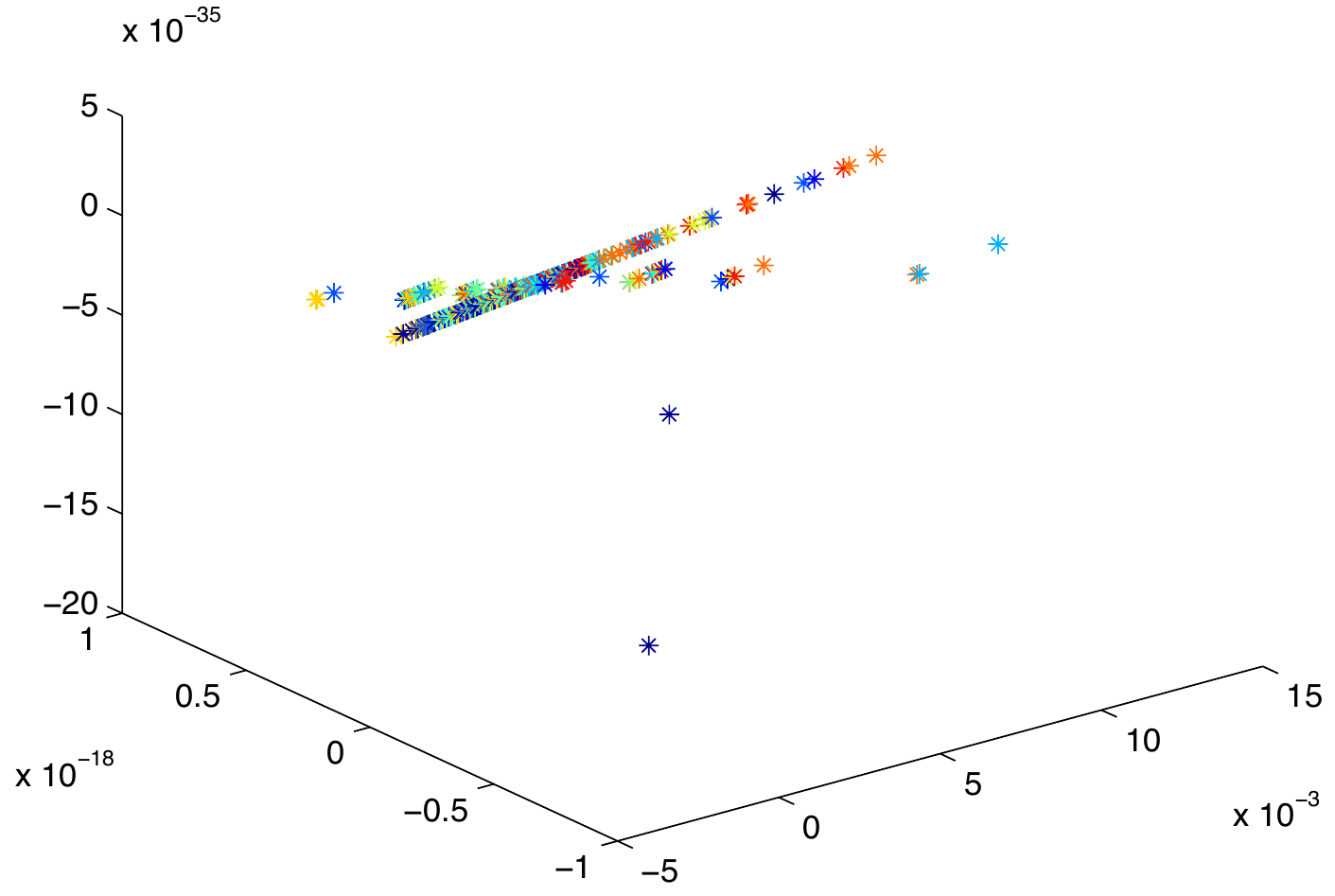}
}
\subfigure[$\nu$, Private]{
\includegraphics[width=3.8cm]{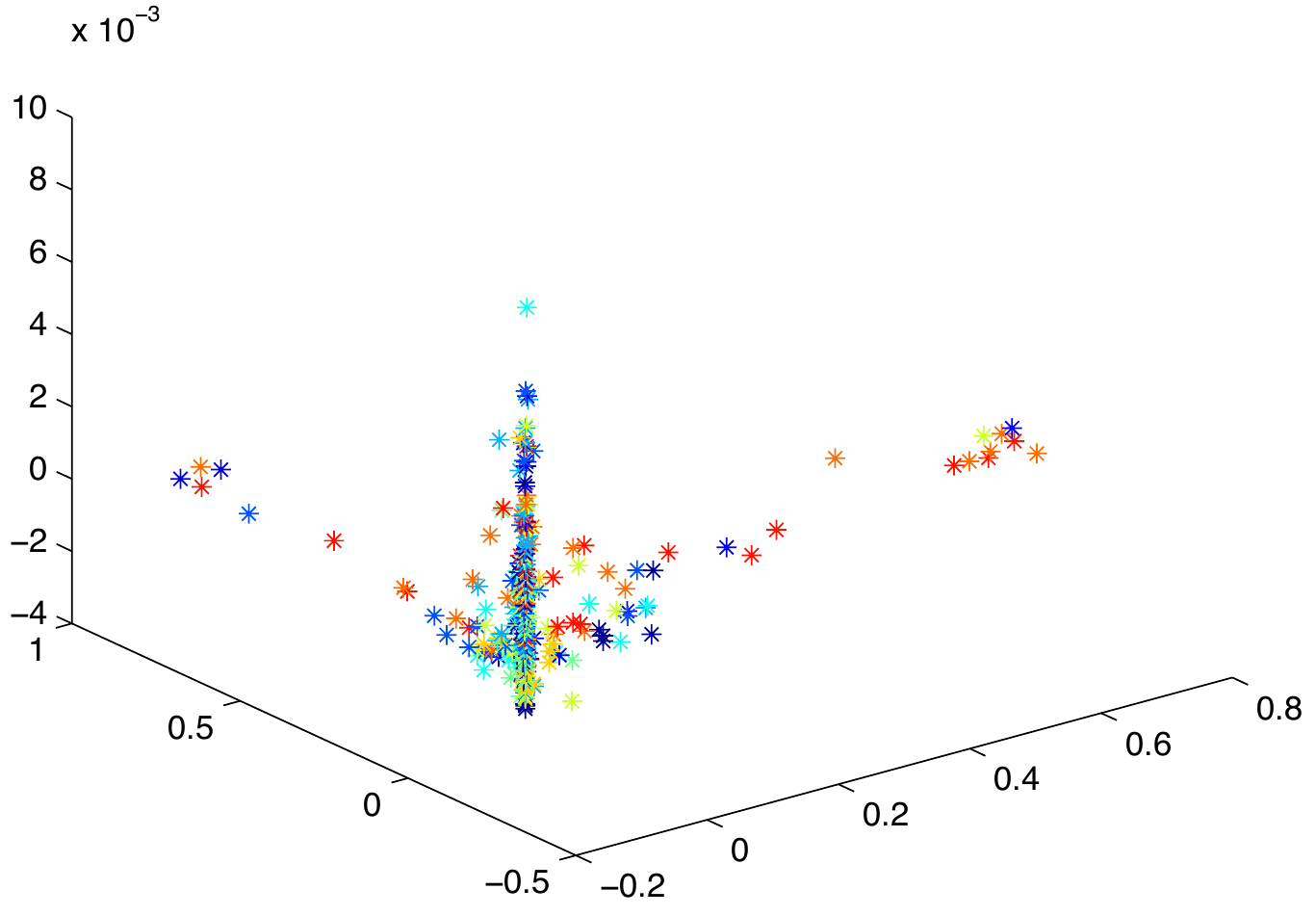}
}
\caption{\scriptsize  Visualization of the shared topic representation ($\theta$) and private topic representations ($\kappa$ and $\nu$) for experiments on the Leeds Butterfly dataset using images paired with their segmentation masks. The documents of different classes are colored differently and the plots show the first three principal components after applying PCA on the per document topic distributions for all the training data. }
\label{fig:doc_topic_imageSeg_Leeds}
\end{figure}

\paragraph{Image and Segmented Image.}
\label{sec:ImageAndSegLeeds}
In this experimental setting, natural images and segmented images are used as two different views for training to demonstrate the multi-modality scenario. The segmented images are used as the first view and the natural images are used as the second view. Since the model is symmetric, the order of the views has no impact on the model.  Figure \ref{fig:hist_imageSegLeeds} shows the histogram of the partition parameter.  It is apparent that the partition parameters of the segmented images are more concentrated around the large values. Thus, the model has learned that the segmented images contain more relevant information. This is consistent with human intuition. Figure \ref{fig:doc_topic_imageSeg_Leeds} shows the topic distribution using shared and private latent representations where the shared topic representations for different classes are naturally separated. Classification performance is summarized in Table \ref{tab:leedsdataset}. { SWB performs better with this dataset than with the LabelMe dataset. The reason for this is probably that the visual words here are less noisy than in LabelMe. } "IS IBTM15"  denotes the performance of testing with only natural images and "IS IBTM15" shows the performance of testing with both natural images and their segmentation. We can see that IBTM performs better than other methods even if only natural images are available for testing. With the segmentation, the performance is almost ideal.

\section{Conclusion}
In this paper, we proposed a different variant of the topic model IBTM with a factored latent representation. It is able to model shared information and private information using different views which has been proven to be beneficial for different computer vision tasks. Experimental results show that IBTM can effectively encode the task-relevant information. Using this representation, the state-of-the-art results are achieved in different experimental scenarios.

In this paper, the focus lay on exploring the concept of factorized representations and the experiments were centered around two view scenarios. In future work,  we plan to evaluate the performance of IBTM by using any number of views and in different scenarios such as cue-integration. In the end, efficient inference algorithms are the key for probabilistic graphic models in general. In this paper, we used variational inference in a batch manner. In the future, more efficient and robust inference algorithms \cite{minka05divergence,hoffman2015structured} can be explored. 

\bibliographystyle{splncs}
\bibliography{ref}

\end{document}